\documentclass{article}

\usepackage{arxiv}

\usepackage[utf8]{inputenc} 
\usepackage[T1]{fontenc}    
\usepackage{hyperref}       
\usepackage{url}            
\usepackage{booktabs}       
\usepackage{amsfonts}       
\usepackage{nicefrac}       
\usepackage{microtype}      

\usepackage{lipsum}         
\usepackage{graphicx}
\usepackage{doi}

\usepackage{adjustbox}
\usepackage{array}
\usepackage{subcaption}
\usepackage{amsmath}
\usepackage{amssymb}
\usepackage{cleveref}       

\title{Merging Reaction to Cognition: A Hybrid Cognitive Strategy for Odour Source Localisation in Natural Environments}

\author{
  Hugo Magalhães, Rui Baptista, and Lino Marques\thanks{Corresponding author: \texttt{lino@isr.uc.pt}} \\
  Institute of Systems and Robotics \\
  Department of Electrical and Computer Engineering \\
  University of Coimbra, 3030-290, Coimbra, Portugal
}

\renewcommand{\shorttitle}{Merging Reaction to Cognition}

\hypersetup{
pdftitle={Merging Reaction to Cognition: A Hybrid Cognitive Strategy for Odour Source Localisation in Natural Environments},
pdfauthor={Hugo Magalhães, Rui Baptista, Lino Marques},
pdfkeywords={odour source localisation, cognitive search, bio-inspired motion, field robots, autonomous decision-making},
}

\begin{document}
\maketitle

\begin{abstract}
Chemical pollutants released into the environment are transported by turbulent flows, generating complex and intermittent plume structures that threaten ecosystems and human health. Rapid localisation of emission sources is critical, and field robots equipped with chemical sensors provide a viable means to perform this task. Nevertheless, inferring source location from sensor readings remains difficult due to sparse detections and the absence of reliable concentration gradients. Existing approaches broadly fall into two paradigms. Bio-inspired strategies rely on reactive behaviours triggered by chemical detections, such as surge–casting, offering efficiency but requiring scenario-specific tuning. Cognitive strategies integrate observations into a probabilistic belief over source location to guide exploration. While more robust, these methods suffer from excessive exploration and strong dependence on belief accuracy. The Fast-Cognitive algorithm alleviated the computational burden of cognitive approaches but preserved their fundamental limitations. Previous Markov chain analysis revealed that source-directed motions occur approximately twice as often following odour detections, indicating that reactive behaviours naturally emerge within cognitive frameworks. This work proposes a hybrid strategy that explicitly incorporates bio-inspired reactivity into belief-dependent motion planning. The method introduces a detection-triggered switching mechanism that formalises transitions between crossflow exploration and source-directed motion, prioritising source proximity over information gain. Behavioural parameters are derived directly from belief metrics, enabling adaptive reactivity without manual tuning. The approach is validated through simulations under three turbulence conditions and field experiments with an autonomous surface vehicle in the Mondego River, Portugal. Results show up to 50\% reduction in traveled distance relative to Fast-Cognitive, with 86\% success rate and 3.2m average localisation error, confirming the effectiveness of combining reactive efficiency with cognitive adaptability.
\end{abstract}

\section*{Acknowledgments}
This work has been partially supported by the Recovery and Resilience Plan (PRR) under Project - Agenda Mobilizadora Sines Nexus - (ref: 7113) and by the Portuguese Foundation for Science and Technology (FCT), Ph.D. studentship SFRH/BD/149527/2019.


\section{Introduction}\label{sec1}

Chemical pollutants released into air or water are transported by turbulent flows and dispersed into complex plume structures that pose significant threats to ecosystems and human health~\cite{marques2022mobile, babuji2023human}. Rapid identification and localisation of emission sources is therefore critical for environmental monitoring and emergency response teams~\cite{rakowska2024searching}. Mobile robots such as Unmanned Aerial Vehicles (UAVs) or Autonomous Surface Vehicles (ASVs) are powerful tools to address these issues with the capability to sample large regions of space and operate in dangerous situations without risking human operators~\cite{kwon2023advancements, liu2024applications, mansfield2024survey}. 

However, locating chemical sources in turbulent plumes remains a central challenge in field robotics. Unstable dispersion fragments chemical signals into sparse, intermittent filaments rendering gradient-based search ineffective~\cite{de2013air, webster2001chemosensory,michaelis2020odor}. Autonomous robots deployed for searching missions must therefore rely on sophisticated Odour Source Localisation (OSL) strategies capable of inferring source location from sporadic and uncertain chemical detections~\cite{yuan2023marine, francis2022gas}.

Two distinct paradigms have emerged to address this challenge. Bio-inspired strategies drawing from insect olfactory behaviours implement reactive behavioral rules, such as surge movements upon chemical detection and casting behaviors when contact is lost~\cite{carde2008navigational, ishida1995odour}. The Silkworm-Moth algorithm exemplifies this approach, achieving efficient source-directed motion through immediate exploitation of chemical cues~\cite{macedo2019comparative, carde2021navigation}. In aquatic environments, fish-inspired approaches have showed effective plume tracking through lateral line sensing and reinforcement learning~\cite{gunnarson2024fish}.This directness and search efficiency represents a key advantage over deliberative methods. However, operating without an internal environment model causes bio-inspired strategies to suffer degraded performance in highly intermittent plumes due to long periods without positive odour detections. Furthermore these methods require scenario-specific parameter tuning, such as surge lengths and casting angles, that must be manually adjusted for each environment~\cite{gaurav2024moth}.

Cognitive strategies offer an alternative information-theoretic approach based on probabilistic reasoning~\cite{vergassola2007}.  These methods maintain a belief distribution over potential source locations,  updated through Bayesian inference from chemical measurements, and select movements that maximize expected information gain~\cite{hutchinson2018information, magalhaes2022evaluating}. Recent advances have incorporated online dispersion simulation to improve source estimation by contrasting real-time gas propagation models against sensor measurements~\cite{ojeda2024robotic}.This approach provides robust adaptability, with search motions emerging from information-seeking tendencies rather than direct reactions to sensor readings.  However, cognitive methods exhibit three significant limitations: high computational  demands from estimating expected information gains at candidate positions; an inherent exploratory tendency that biases the search toward uncertainty reduction rather than direct source approach~\cite{jing2021recent}; and complete dependency on belief accuracy, causing mission failure when estimation diverges. The Fast-Cognitive algorithm addressed the computational cost by replicating informative movement patterns through belief-derived search behaviours, avoiding the need for future information gain estimation~\cite{magalhaes2025fastcog}. However, by replicating informative motions, it retained the exploratory tendency and belief dependency inherent to cognitive approaches.

The directness and efficiency of bio-inspired reactive motions offer a potential solution for cognitive over-exploration, while belief-driven parameterization avoids manual adjustment of search behaviours without sacrificing reliability. These complementary advantages motivate a hybrid approach, with recent work revealing that reactive-probabilistic switching mechanisms can improve search efficiency in turbulent conditions~\cite{bourne2019coordinated, luong2024reactive,wang2024exploration}.Support for this integration emerges from Markov chain analysis of cognitive search behaviors~\cite{magalhaes2025motion}, which revealed that cognitive strategies naturally produce detection-dependent behavioral transitions strikingly similar to bio-inspired surge-casting patterns. Without odour contact, agents exhibited predominantly crossflow exploratory behaviors (69.7\%). Upon positive chemical detection, source-directed motions increased from 23.6\% to 41.7\%, showing a transition analogous to surge responses in insect navigation. This spontaneous emergence of bio-inspired patterns from belief-driven decisions reveals a fundamental convergence between paradigms, suggesting that reactivity can be explicitly integrated into cognitive frameworks without sacrificing adaptive parameterization.

Following this insight, this work proposes a hybrid reactive-cognitive OSL strategy, denoted as Hybrid Fast-Cognitive, that implements detection-triggered belief-based behavioral switching. Upon chemical contact, motion immediately shifts from exploratory crossflow patterns to source-directed behaviors, drawing inspiration from the transitions identified through Markov analysis. This reduces over-exploration by prioritizing source approach over uncertainty reduction when chemical evidence is present. Simultaneously, reactive responses provide source-directed motion independent of belief accuracy, reducing vulnerability to estimation divergence. Motion parameters remain dynamically determined from belief metrics, enabling adaptive reactivity without scenario-specific tuning.

The proposed strategy is validated through simulated experiments across three turbulence scenarios and field trials in the Mondego River, in Portugal. Simulations shows up to 26\% reduction in overall traveled distance compared to Fast-Cognitive. Under highly turbulent conditions, where cognitive over-exploration and belief divergence are most pronounced, the hybrid strategy maintained approximately 90\% success rate while other methods dropped to 60\%. Field experiments employed an autonomous surface vehicle equipped with a conductivity probe to locate a salt-water plume released from a stationary floating platform located upstream, creating realistic dispersion conditions. The hybrid strategy achieved 86\% success rate with 3.17m average localization error and reduced travel distance by more than 50\% compared to Fast-Cognitive field experiments, confirming that belief-based reactivity substantially improves search efficiency in real-world conditions.

The remainder of this paper is organized as follows. Section~II formulates the OSL problem and summarizes the belief estimation framework. Section~III describes bio-inspired strategies and the original Fast-Cognitive approach. Section~IV presents the hybrid strategy, detailing the detection-triggered switching mechanism and belief-driven parameterization. Section~V reports simulation and field experimental results. Section~VI concludes with discussion of limitations and future directions.

\section{Problem formulation}\label{sec2}

Consider a mobile agent at positions $\mathbf{p} = (x,y) \in \mathbb{R}^2$ sensing chemical concentrations $c(\mathbf{p}, t)$ in a bounded workspace $\Omega$ with dominant flow direction $\vec{u}$. The agent must locate an unknown source at $(x_s, y_s)$ releasing contaminant at rate $Q$. 
This work adopts the Gaussian plume dispersion model and particle filter-based Bayesian inference framework established previously in~\cite{magalhaes2025motion}, which maintains a probabilistic belief $b_t$ over the source and environmental parameters $\theta = [x_s, y_s, Q, D_y, D_z]$, where $D_y$ represents the dispersion coefficient in the horizontal crossflow direction ($y$) and $D_z$ represents the dispersion along the depth ($z$).

Building upon Fast-Cognitive, the hybrid OSL strategy proposed in this work introduces chemical detection $c_t$ as an explicit input to the decision process which generates actions $a_t$ under the search behaviours $\pi$:
\begin{equation}
    a_t = g(c_t, b_t; \pi)
\end{equation}

The source is considered found when the searching agent moves over the true source location or when it reaches the vicinity of the estimated source and the uncertainty $U_t$ of the belief drops below a specified threshold $\theta_1$.

\section{Methods}\label{sec3}

\subsection{Bio-Inspired Search Strategies}

Bio-inspired OSL strategies draw from different animals olfactory strategies, in particular from insects, such as moths that locate pheromone sources through reactive behaviors triggered by chemical detection~\cite{carde2008navigational}. These strategies operate without internal environment models, instead relying on immediate sensory feedback to guide motion. The Silkworm-Moth algorithm~\cite{ishida1995odour} exemplifies this approach through three discrete behaviors: surge movements upflow upon chemical detection, casting motions perpendicular to flow when contact is lost and increasing spirals to resample previous detection zones. Parameters including surge length $l_s$, cast angle $\alpha_c$, and cast length $l_c$ must be configured in advanced for each deployment scenario~\cite{gaurav2024moth}. 

\subsection{Fast-Cognitive Approach}

The Fast-Cognitive algorithm~\cite{magalhaes2025fastcog} achieves computational efficiency while maintaining cognitive reliability by replicating informative movement patterns through belief-derived behavioral rules. The approach combines exploitative and exploratory motion vectors,$\vec{v}_e$ and $\vec{v}_x$ respectively, weighted by belief uncertainty:
\begin{equation}
    \vec{v}_t = (1 - \alpha) \cdot \vec{v}_e + \alpha \cdot \vec{v}_x
\end{equation}
where $\alpha = U_t/U_{t=1}$ adapts exploration-exploitation balance based on belief uncertainty $U_t$. The exploitative vector $\vec{v}_e$ directs the agent toward the estimated source with magnitude proportional to the estimated plume width $3\sigma_{y,t}$, while the exploratory vector $\vec{v}_x$ incorporates an $80\deg$ angular deviation (average value of movement deviations relative to the source direction under high belief uncertainty, observed in the motion patterns study in~\cite{magalhaes2025motion}) from the source direction with magnitude determined by the farthest estimated plume boundary. Two additional behaviors enhance robustness. A memory recovery behavior activates when the agent approaches the estimated source with high uncertainty, redirecting search toward previous odour contact locations. A source declaration behavior performs an elliptical trajectory around the estimated position when uncertainty falls below a specified threshold, refining the belief before final localization. For complete algorithmic details, see~\cite{magalhaes2025fastcog}.

\section{Hybrid Fast-Cognitive Strategy}\label{sec4}

The proposed hybrid strategy introduces detection-triggered behavioral switching into the Fast-Cognitive framework, implementing the reactive-cognitive integration formulated in Section~2.

\subsection{Detection-Triggered Behavioral Switching}

The core innovation is a binary switching mechanism based on instantaneous chemical detection status $c_t$:
\begin{equation}
    \vec{v}_t = 
    \begin{cases}
        \vec{v}_e & \text{if } c_t \geq \tau_d \\
        \vec{v}_x & \text{otherwise}
    \end{cases}
\end{equation}
where $\tau_d$ is the detection threshold. Upon positive detection, motion shifts to source-directed exploitative behavior $\vec{v}_e$. When contact is lost, crossflow exploratory behavior $\vec{v}_x$ searches for plume re-acquisition. This implements the detection-dependent behavioral transitions identified through Markov analysis~\cite{magalhaes2025motion}, explicitly realizing the natural convergence between cognitive and bio-inspired patterns. Unlike the original Fast-Cognitive formulation, which weights $\vec{v}_e$ and $\vec{v}_x$ continuously based on belief uncertainty $\alpha$, the hybrid approach uses detection status to discretely select behavioral modes. This prioritizes source approach over uncertainty reduction when chemical evidence is present, reducing the exploratory tendency and belief dependency inherent to information-seeking strategies.

\subsection{Belief-Driven Parameterization}

Unlike bio-inspired strategies requiring prior parameter tuning, search behaviours derive their parameters dynamically from belief metrics, maintaining the adaptive parameterization of Fast-Cognitive.

\textbf{Exploitative behavior} ($c_t \geq \tau_d$): Upon chemical detection, the source-directed vector $\vec{v}_e$ guides the agent toward the estimated source:
\begin{equation}
    \vec{v}_e = l_e \cdot (\cos(\alpha_e), \sin(\alpha_e))
\end{equation}
where $\alpha_e = \tan^{-1}(y_s - y, x_s - x)$ represents the angle between the actual agent position and the predicted source location. The magnitude $l_e \propto \sigma_{y,t}$ scales with estimated plume width, similar to Fast-Cognitive by following the findings of informative motion patterns from~\cite{magalhaes2025motion}.

\textbf{Exploratory behavior} ($c_t < \tau_d$): When contact is lost, the crossflow vector $\vec{v}_x$ searches for plume re-acquisition:
\begin{equation}
    \vec{v}_x = l_x \cdot (\cos(\alpha_x), \sin(\alpha_x))
\end{equation}
where $\alpha_x = \alpha_e \pm \gamma$ incorporates an angular deviation from the source direction $\gamma$, and $l_x$ is determined by the farthest estimated plume boundary distance, targeting high information gain regions~\cite{magalhaes2025entropy}. The sign alternation ensures systematic coverage of both plume sides.

\begin{figure*}[tbp]
      \centering
      \includegraphics[width=0.7\linewidth]{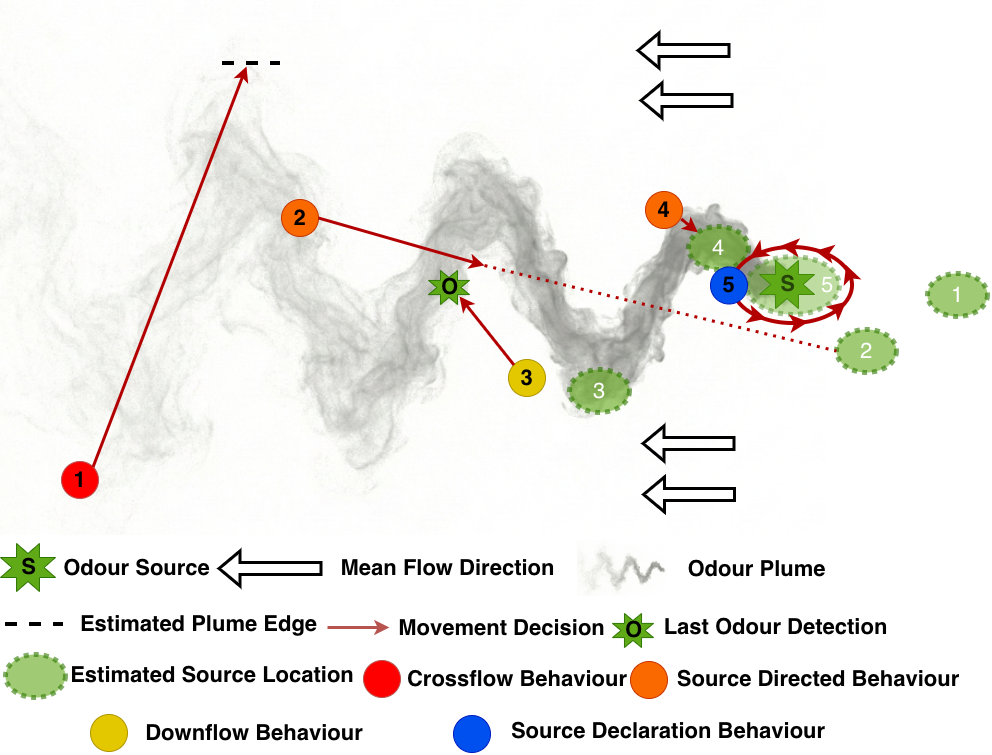}
      \caption{Hybrid Fast-Cognitive decisions, showing adaptive adjustment of motion parameters and behaviour switching based on odour encounters.}
      \label{fig:dec_making}
   \end{figure*}

Figure~\ref{fig:dec_making} illustrates how these parameters adapt to belief state and react to odour encounters. Under high uncertainty, exploratory behavior dominates with wider angular deviations and longer step lengths. As certainty increases, motion becomes increasingly source-directed with shorter, more focused movements.

This formulation preserves the belief-derived parameterization that enables Fast-Cognitive to adapt to varying environmental conditions, while the detection-triggered mode selection introduces the directness characteristic of bio-inspired approaches.

\subsection{Supporting Behaviors}

Three additional behaviors from Fast-Cognitive complement the hybrid movement decision to enhance robustness:

\textbf{Plume Search:} Before initial chemical contact, the agent executes zig-zag patterns moving upflow to locate the active plume region~\cite{marjovi2014optimal}. Once the target substance is detected, the strategy transitions to hybrid OSL decision-making;

\textbf{Memory Recovery:} When the agent approaches the estimated source ($\sqrt{(x - x_s)^2 + (y - y_s)^2}<\theta_2$) with high uncertainty ($U_t > \theta_1$), indicating potential belief divergence, the strategy redirects search toward previous odour contact locations stored in memory. This provides a recovery mechanism when belief estimation fails;

\textbf{Source Declaration:} When uncertainty $U_t$ falls below threshold $\theta_1$ and the agent is near the estimated position, an elliptical trajectory around the estimated source and aligned with the dominant fluid flow direction $u_d$ is performed to refine the belief before final localization. The horizontal and vertical radii of the ellipse the ellipse, $r_h$ and $r_v$, are scaled according to the uncertainty in the estimated source coordinates $r_v \propto \sqrt{\theta_{y_s}}, \quad r_h \propto \sqrt{\theta_{x_s}}$.

\subsection{Behavior Tree Implementation}

\begin{figure*}[tbp]
      \centering
      \includegraphics[width=1.0\linewidth]{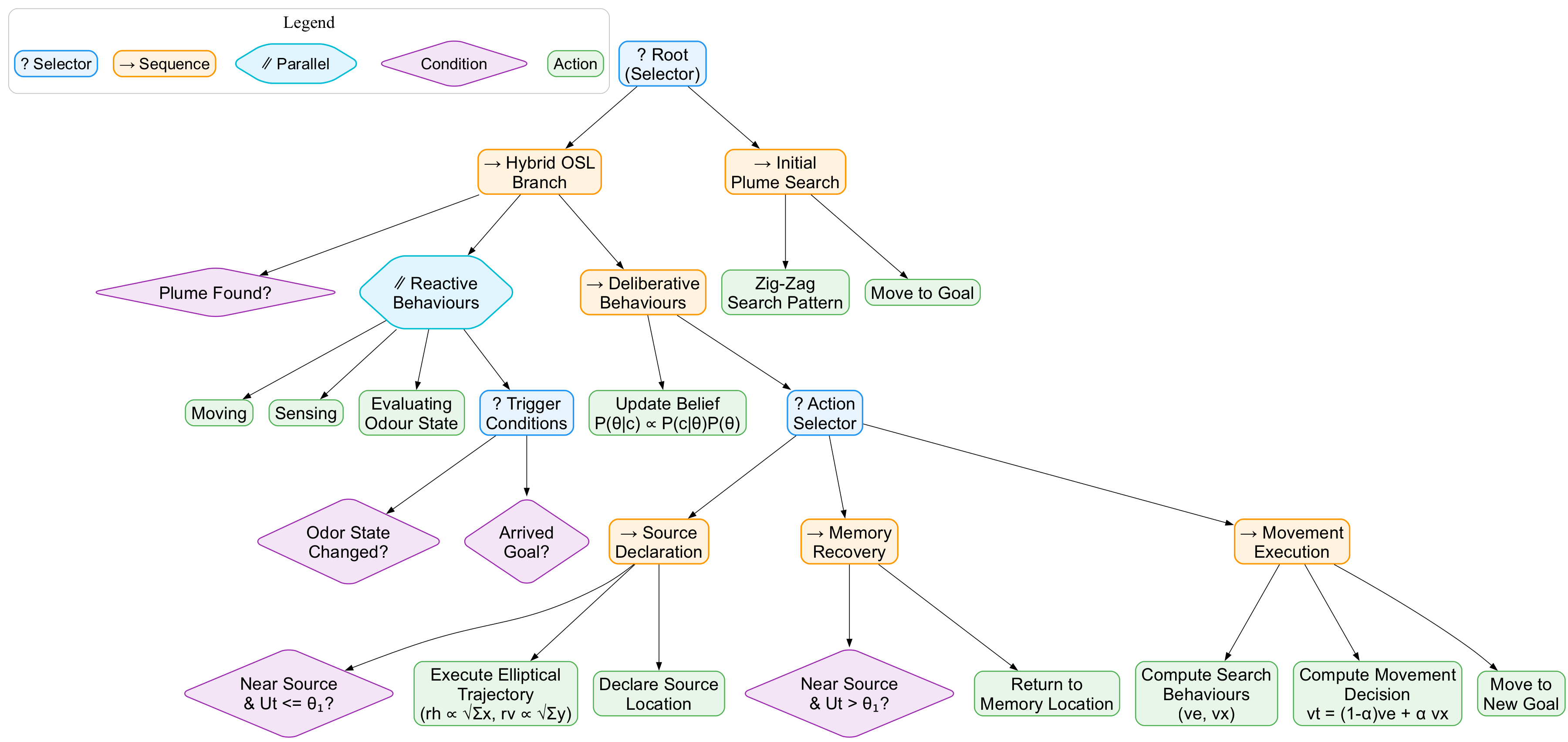}
      \caption{Hybrid Fast-Cognitive Behaviour Tree.}
      \label{fig:decmakingdiagram}
   \end{figure*}

The hybrid strategy is implemented using a Behavior Tree (BT) framework~\cite{gugliermo2024evaluating, ogren2022behavior}, following established principles for autonomous robot architectures~\cite{ingrand2017deliberation}. BTs are particularly well-suited for this application because the tick-based execution model enables real-time behavioral switching in response to sensory events, while the hierarchical structure allows modularity and a natural integration of reactive responses with deliberative belief-driven planning. The BT framework is composed by three main primary control node types that govern the execution flow: (1) Selector nodes (denoted by symbol $?$) prioritize between alternative behaviors by triggering children nodes until one succeeds, (2) Sequence nodes (represented by symbol $\rightarrow$) activate children nodes in order, failing if any child fails, and (3) Parallel nodes (symbol $\parallel$) run multiple children nodes concurrently, enabling simultaneous operations such as sensing, reasoning, and acting. Furthermore, the modular tree structure facilitates the future incorporation of learning methods through direct tree evolution or policy optimization~\cite{scheide2021behavior}. Figure~\ref{fig:decmakingdiagram} illustrates the complete approach.

The BT employs a root selector prioritizing between Initial Plume Search and the Hybrid OSL Branch. Once the plume is detected, the system transitions from zig-zag search patterns to hybrid decision-making. A Reactive Decision Branch executes parallel processes for Moving, Sensing, Evaluating Odour State, and monitoring Trigger Conditions. This enables real-time behavioral switching. When the agent detects chemicals, it immediately transitions to exploitative source-directed behavior. When contact is lost, it switches to exploratory crossflow patterns. The Deliberative Branch computes belief-derived motion parameters that feed into the behavioral mode that is active. This architecture provides the fast response characteristic of bio-inspired approaches while retaining the adaptive parameterization of cognitive methods, achieving a complete hybrid OSL strategy.

\section{Experimental Evaluation}\label{sec5}

\subsection{Simulations}

\begin{figure*}[tbp]
    \centering
    \begin{subfigure}{0.32\textwidth}
        \centering
        \includegraphics[width=\linewidth]{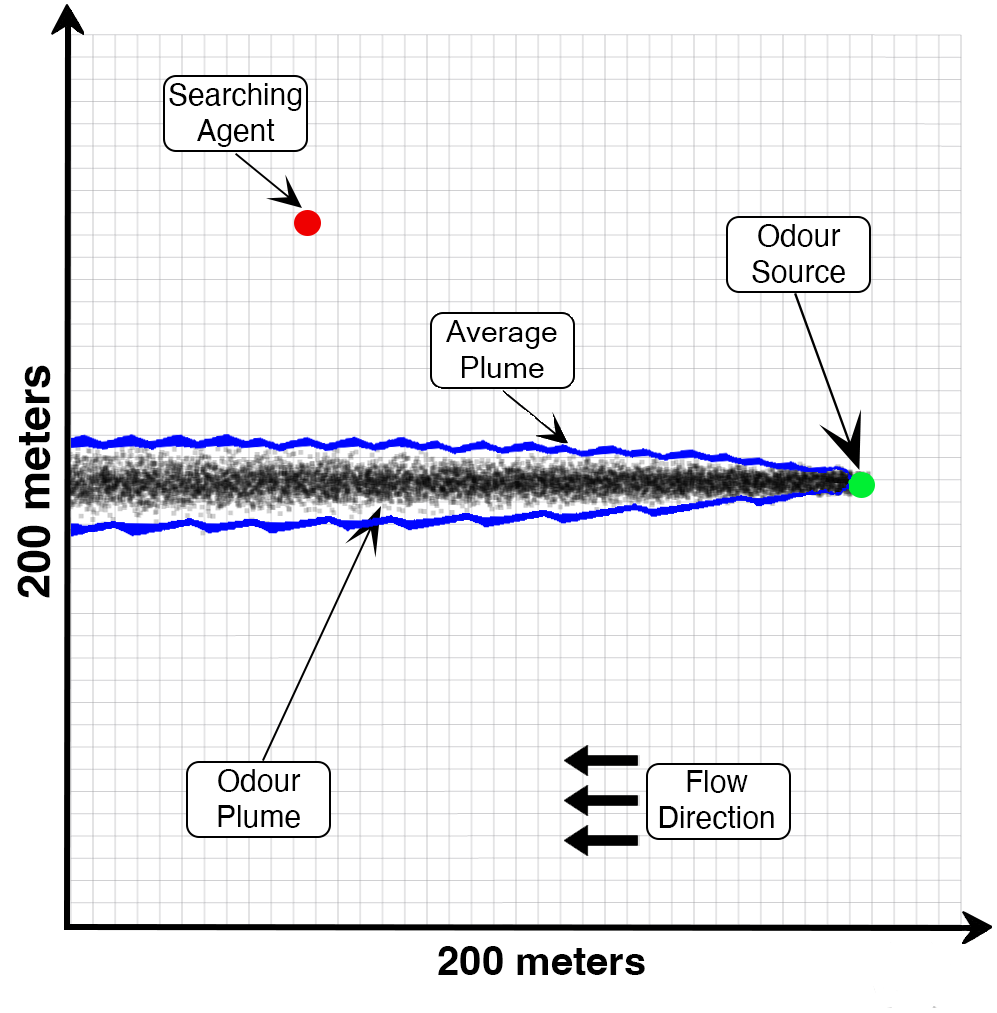}
        \caption{}
        \label{fig:sima}
    \end{subfigure}
    \hfill
    \begin{subfigure}{0.32\textwidth}
        \centering
        \includegraphics[width=\linewidth]{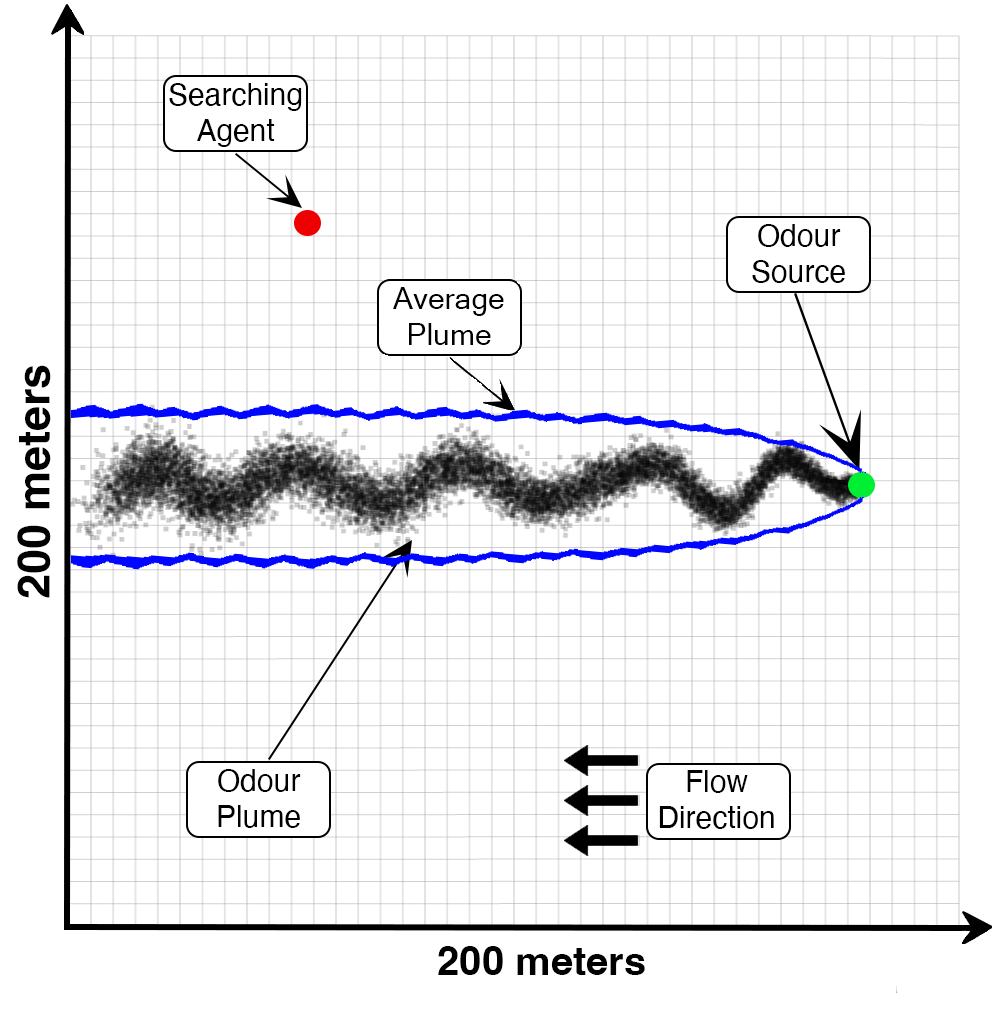}
        \caption{}
        \label{fig:simb}
    \end{subfigure}
    \hfill
    \begin{subfigure}{0.32\textwidth}
        \centering
        \includegraphics[width=\linewidth]{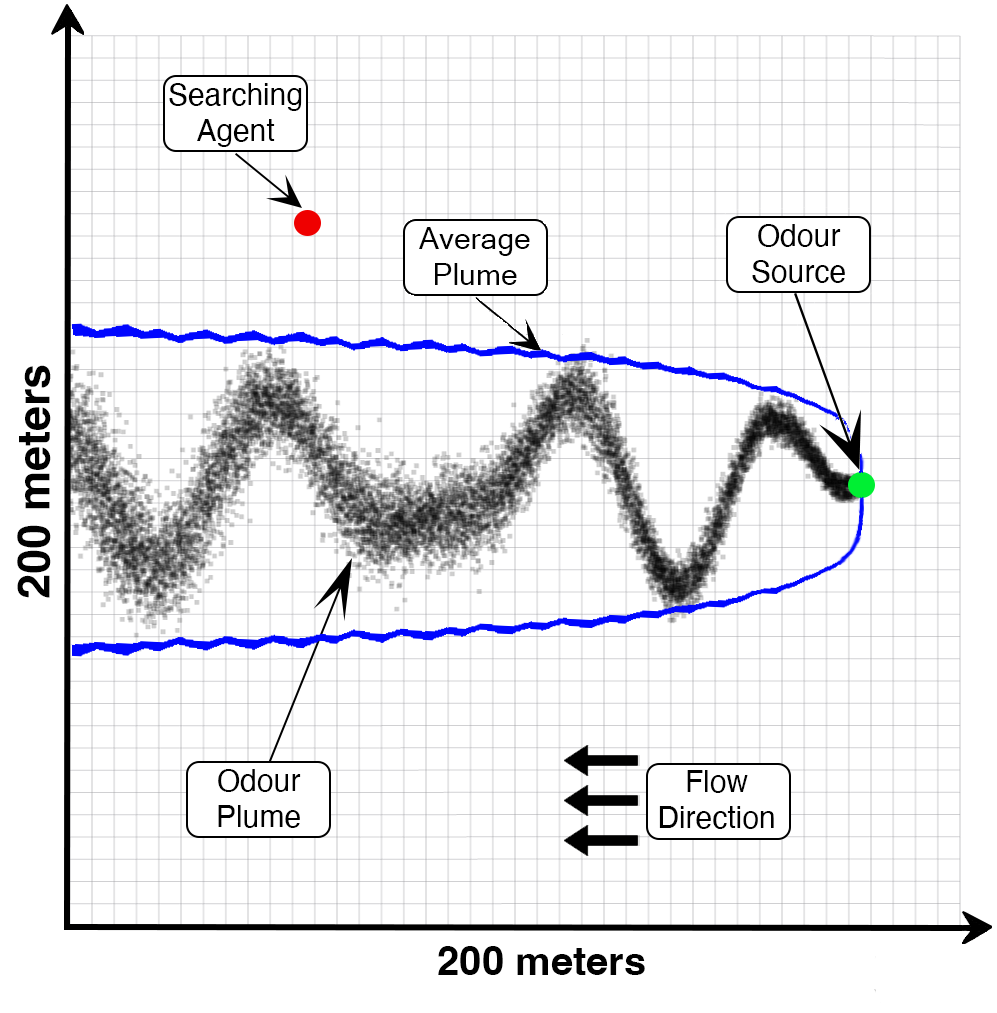}
        \caption{}
        \label{fig:simc}
    \end{subfigure}
    \caption{Simulated environments (a) Scenario 1 with a stable plume (b) Scenario 2 with a moderately unstable plume, (c) Scenario 3 with highly unstable conditions.}
    \label{fig:simulated_plumes}
\end{figure*}

The performance of the proposed algorithm was evaluated in three simulated environmental scenarios with increasing turbulence levels as in previous Fast-Cognitive evaluation, namely: stable flow generating an undistorted plume (S1, Figure~\ref{fig:simulated_plumes}a); moderate transversal oscillations producing a meandering plume (S2, Figure~\ref{fig:simulated_plumes}b); and high turbulence creating a highly distorted, intermittent plume (S3, Figure~\ref{fig:simulated_plumes}c). These simulations use the filament-based model for plume dispersion proposed by Farrell in~\cite{farrell2002}. The hybrid strategy (represented by H) is compared against Fast-Cognitive (represented by the letter F) across 200 experiments per scenario, with agents starting approximately 80m downflow from the source. The source intensity $Q$ and the detection threshold $\tau_d$ assure that the agents start searching outside the active plume area.  The movement speed of the agents is programmed as $0.5 m/s$. The experimental parameters are shown in Table~\ref{tab:oslparameters}, where $\sigma_m$ represents the uncertainty parameter for the likelihood function in the Bayesian Inference, and $u_s$ and $u_d$ denote the measurements of fluid speed and direction respectively.

\begin{table}[tbp]
\caption{Parameters used in the simulations.}
\label{tab:oslparameters}
\centering
\begin{tabular}{|ll|ll|}
\hline
\multicolumn{4}{|c|}{Search Parameters} \\ 
\hline
\multicolumn{2}{|c|}{Inference and Decision} & \multicolumn{2}{c|}{Prior} \\ 
\hline
$N$ & 5000 & $x_s$ & $\mathcal{U}(x',x'+150)$ m \\
Resampling & 50\% & $y_s$ & $\mathcal{U}(y',y'+30)$ m \\
$\sigma_m$ & 10 & $Q$ & $\mathcal{U}(0.01, 1000)$ g/s \\
$\tau_d$ & 1.0 g/m$^3$ & $D_y$, $D_z$ & $\mathcal{U}(0.001, 0.2)$ m$^2$/s \\
 $\theta_1$ & 5 m & $u_s$ & 1 m/s \\
 $\theta_2$ &2 m & $u_d$ & $\pi$ rad \\
\hline
\end{tabular}
\end{table}

The comparison between Fast-Cognitive and Hybrid Fast-Cognitive is conducted through two complementary analyses. Statistical analysis evaluates success rate, traveled distance, distance ratio, and computational and search times across all experiments. An experiment is considered successful if the agent localizes the source within 5 meters.Trajectory analysis examines individual and aggregated search paths to characterize motion dynamics.

\subsubsection{Statistical Analysis}

\begin{figure*}[tbp]
    \centering
    \begin{subfigure}{0.48\textwidth}
        \centering
        \includegraphics[width=\linewidth]{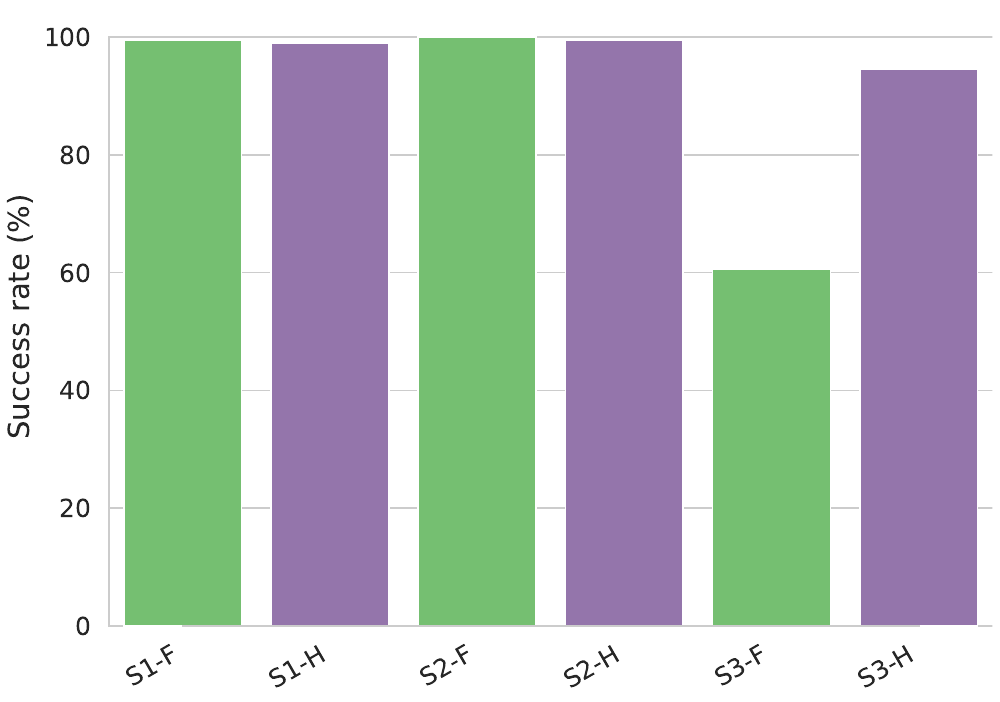}
        \caption{}
        \label{fig:stat_dec_it}
    \end{subfigure}
    \hfill
    \begin{subfigure}{0.48\textwidth}
        \centering
        \includegraphics[width=\linewidth]{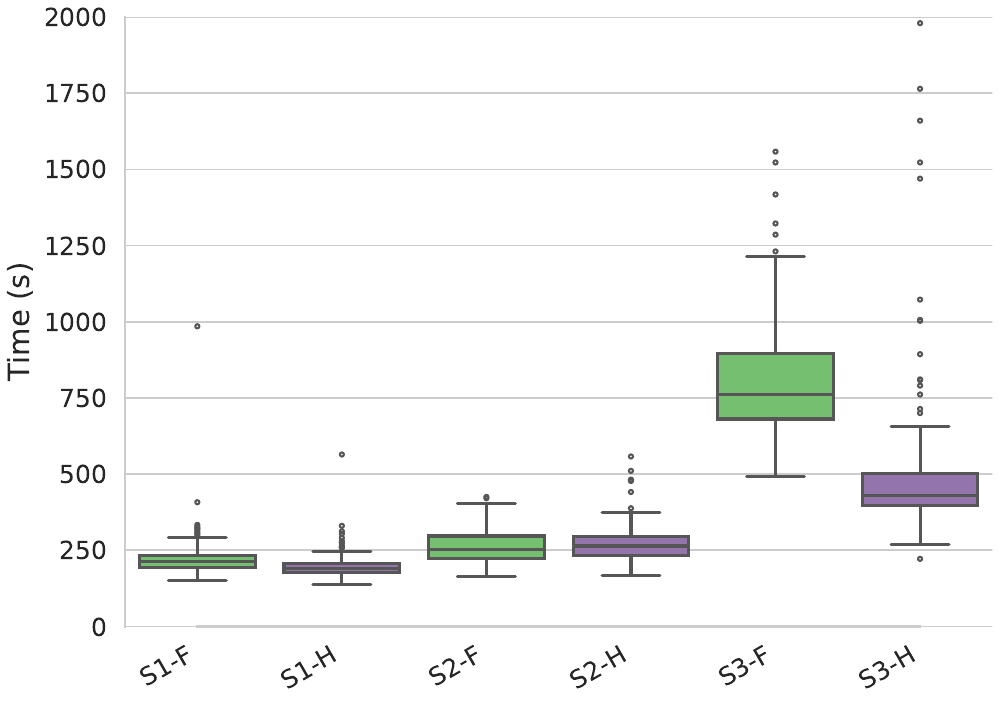}
        \caption{}
        \label{fig:stat_dec_time}
    \end{subfigure}\\
       \begin{subfigure}{0.48\textwidth}
        \centering
        \includegraphics[width=\linewidth]{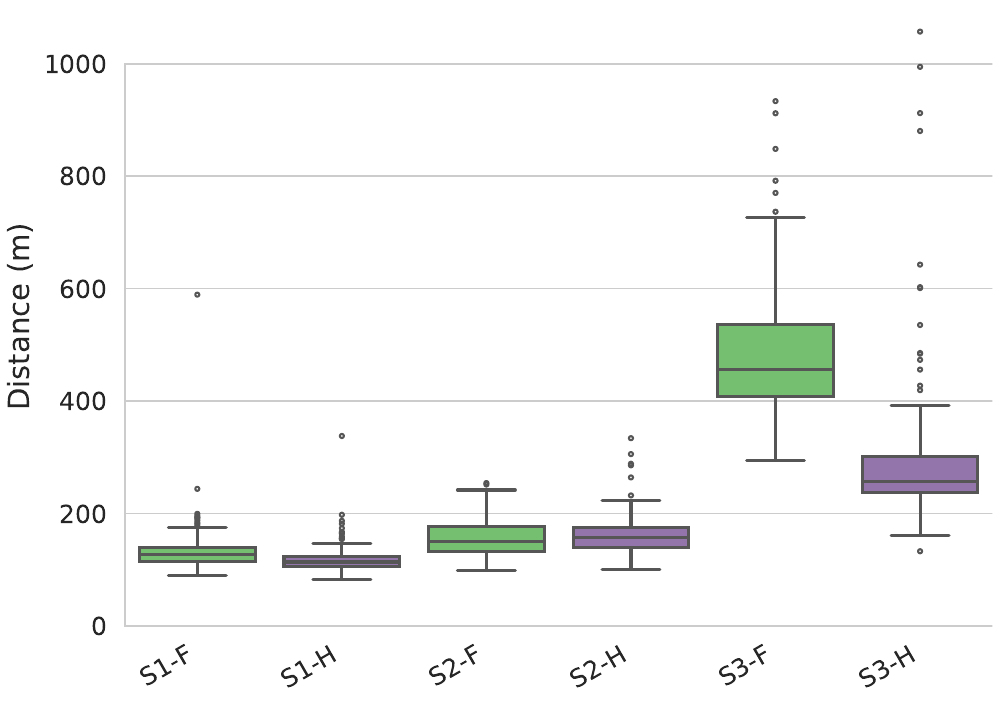}
        \caption{}
        \label{fig:stat_moving_time}
    \end{subfigure}
    \hfill
    \begin{subfigure}{0.48\textwidth}
        \centering
        \includegraphics[width=\linewidth]{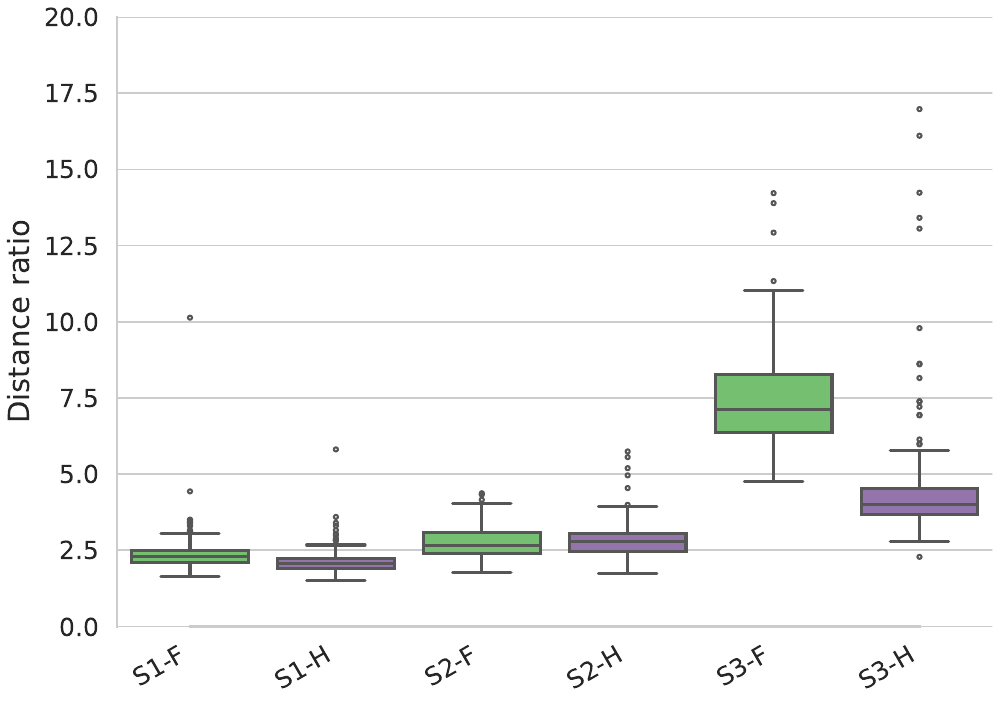}
        \caption{}
        \label{fig:stat_total_search}
    \end{subfigure}
    \caption{(a) success rates, (b) total search time per experiment, (c) accumulated traveled distance per experiment and (d) distance ratio between shortest path (from first odour detection) to the source and traveled distance.}
    \label{fig:sim_stats}
\end{figure*}

Figure~\ref{fig:sim_stats} presents detailed performance comparisons between F and H across all three scenarios.

\textbf{Success Rate:} Both methods achieve high reliability in simpler scenarios (S1, S2), with success rates between 95\% and 100\% (Figure~\ref{fig:sim_stats}a). Under highly turbulent conditions (S3), significant differences emerge: the hybrid strategy maintained approximately 93\% success rate while Fast-Cognitive dropped to 61\%. This improvement stems from reduced belief dependency, where reactive responses provide source-directed motion even when belief estimation diverges, enabling recovery from situations that cause pure cognitive methods to fail.

\textbf{Total Search Time:} Combining movement and computational times, the hybrid strategy reduces total search time by approximately 26\% compared to Fast-Cognitive (Figure~\ref{fig:sim_stats}b). This improvement results primarily from reduced travel distances, as the hybrid mechanism introduces no significant computational overhead.  The most notable improvements are observed in S3 with a reduction of approximately 60\% due to plume complexity, with the remaining two scenarios originating similar values.

\textbf{Traveled Distance:} The hybrid approach reduces traveled distance by approximately 26\% on average compared to Fast-Cognitive (Figure~\ref{fig:sim_stats}c). In S1 and S2, both methods perform similarly due to narrow, less distorted plume shapes. Pronounced improvements occur in S3, where the hybrid strategy requires substantially shorter paths, whereas F require extended distances to reduce belief uncertainty. This efficiency gain reflects the reduced exploratory tendency, where upon chemical detection, the agent prioritizes source approach over uncertainty reduction.

\textbf{Distance Ratio:} The distance ratio quantifies search directness by comparing traveled distance to straight-line distance from first detection to source (Figure~\ref{fig:sim_stats}d). Fast-Cognitive exhibits an average ratio of 4.21, reflecting substantial exploration that increases in more turbulent scenarios to values closer to 7.5. The hybrid strategy achieves 3.19 on average, representing an overall reduction of 24\%, and a value of 4.5 in S3 which translates into a reduction of 60\%.  These results confirm that the combination of reaction and cognition effectively reduces over-exploration while maintaining higher success rates.

Table~\ref{tab:sim_summary} summarizes performance metrics aggregated across all scenarios, confirming consistent improvements in reliability, efficiency, and search directness.

\begin{table}[tbp]
\centering
\caption{Summary of simulation performance metrics (mean $\pm$ std)}
\label{tab:sim_summary}
\begin{tabular}{lcc}
\toprule
\textbf{Metric} & \textbf{F} & \textbf{H} \\
\midrule
Success Rate (\%) & $86.67 \pm 18.50$ & $97.67 \pm 2.25$ \\
Total Search Time (s) & $432.76 \pm 292.38$ & $319.32 \pm 185.33$ \\
Traveled Distance (m) & $258.92 \pm 175.15$ & $191.15 \pm 111.05$ \\
Distance Ratio & $4.21 \pm 2.53$ & $3.19 \pm 1.82$ \\
\bottomrule
\end{tabular}
\end{table}

\subsubsection{Trajectory Analysis}

\begin{figure*}[tbp]
    \centering
    \begin{subfigure}{0.315\textwidth}
        \centering
        \includegraphics[width=\linewidth]{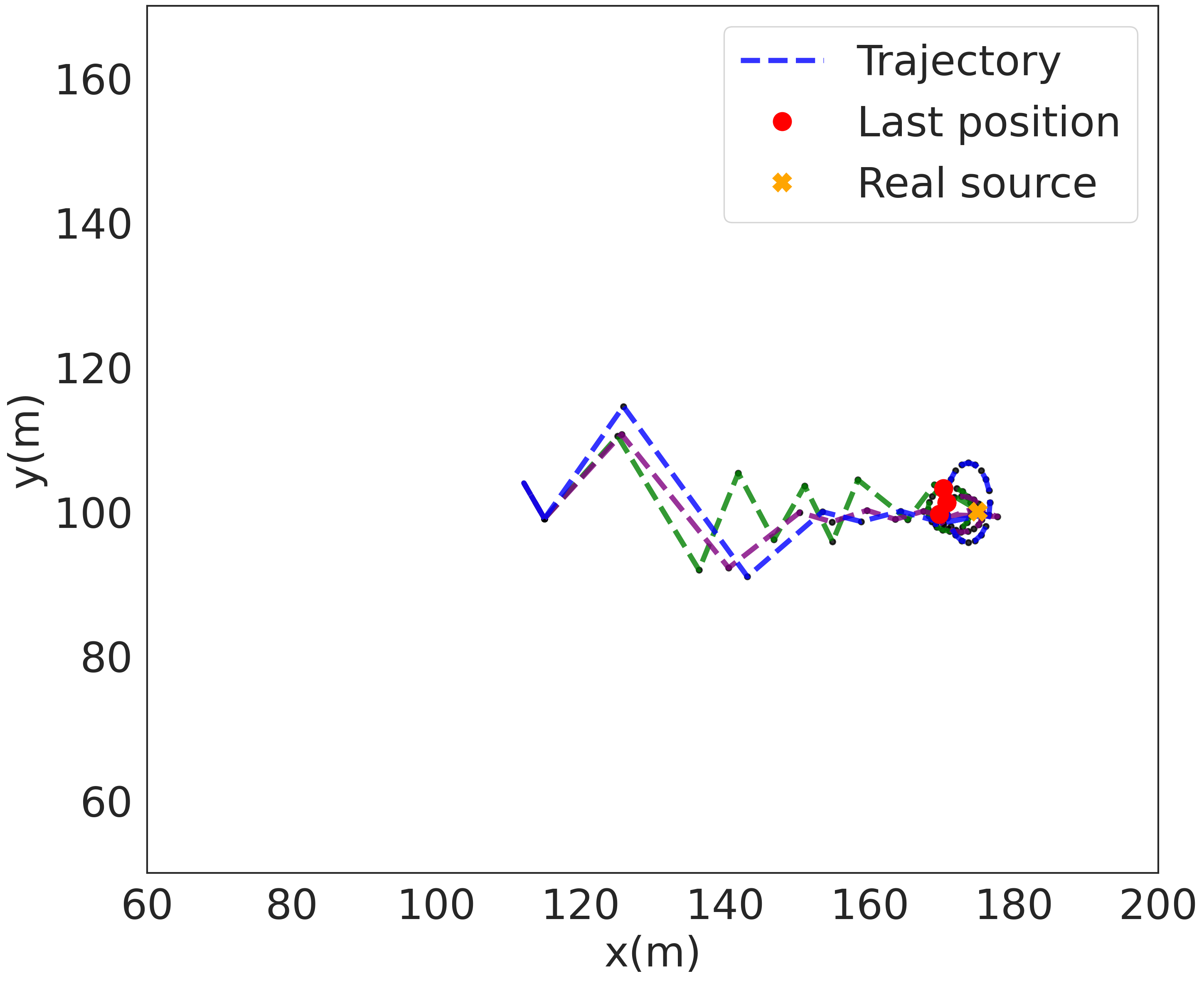}
        \caption{}
        \label{fig:t1}
    \end{subfigure}
    \hfill
    \begin{subfigure}{0.315\textwidth}
        \centering
        \includegraphics[width=\linewidth]{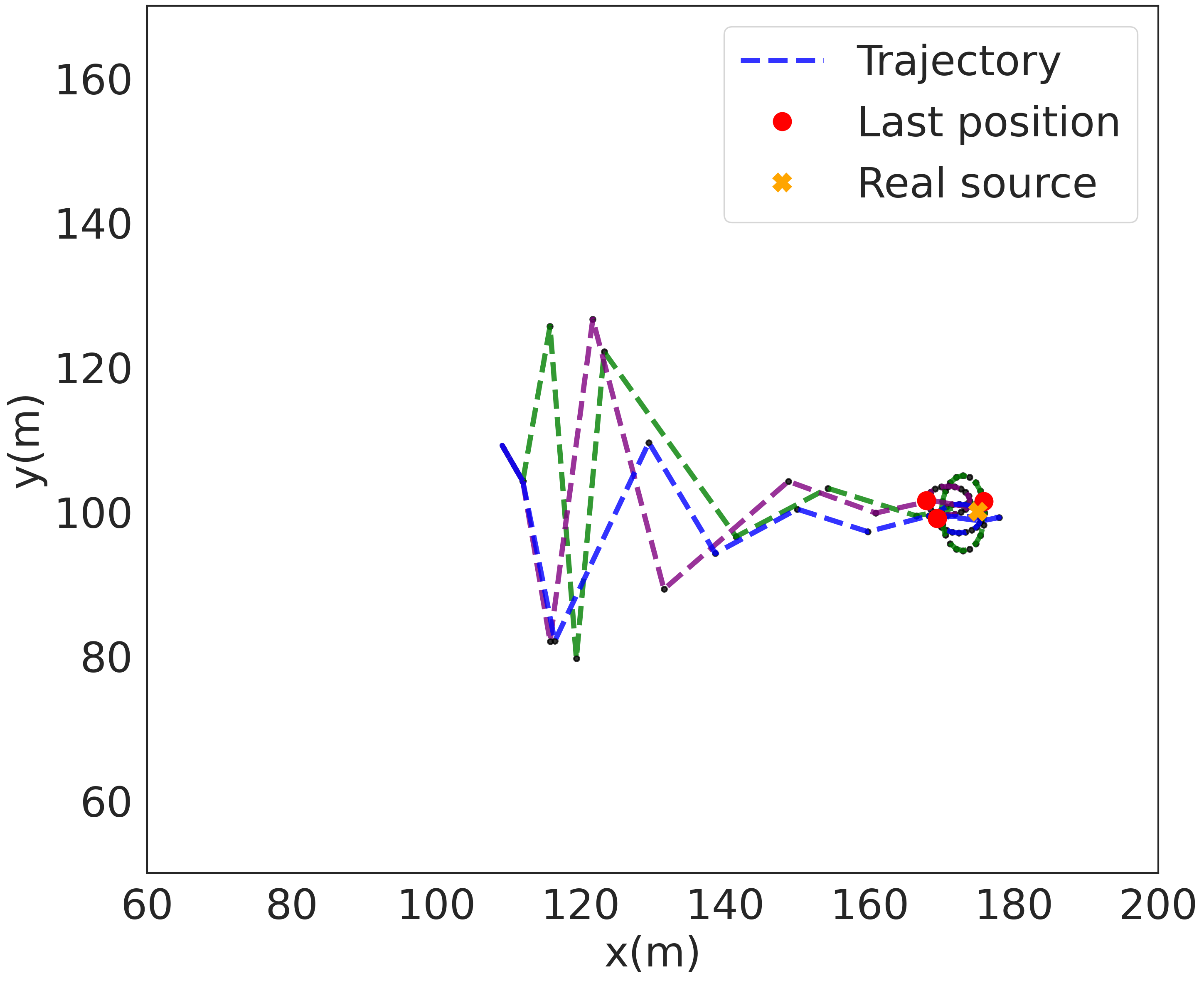}
        \caption{}
        \label{fig:t2}
    \end{subfigure}
    \hfill
    \begin{subfigure}{0.315\textwidth}
        \centering
        \includegraphics[width=\linewidth]{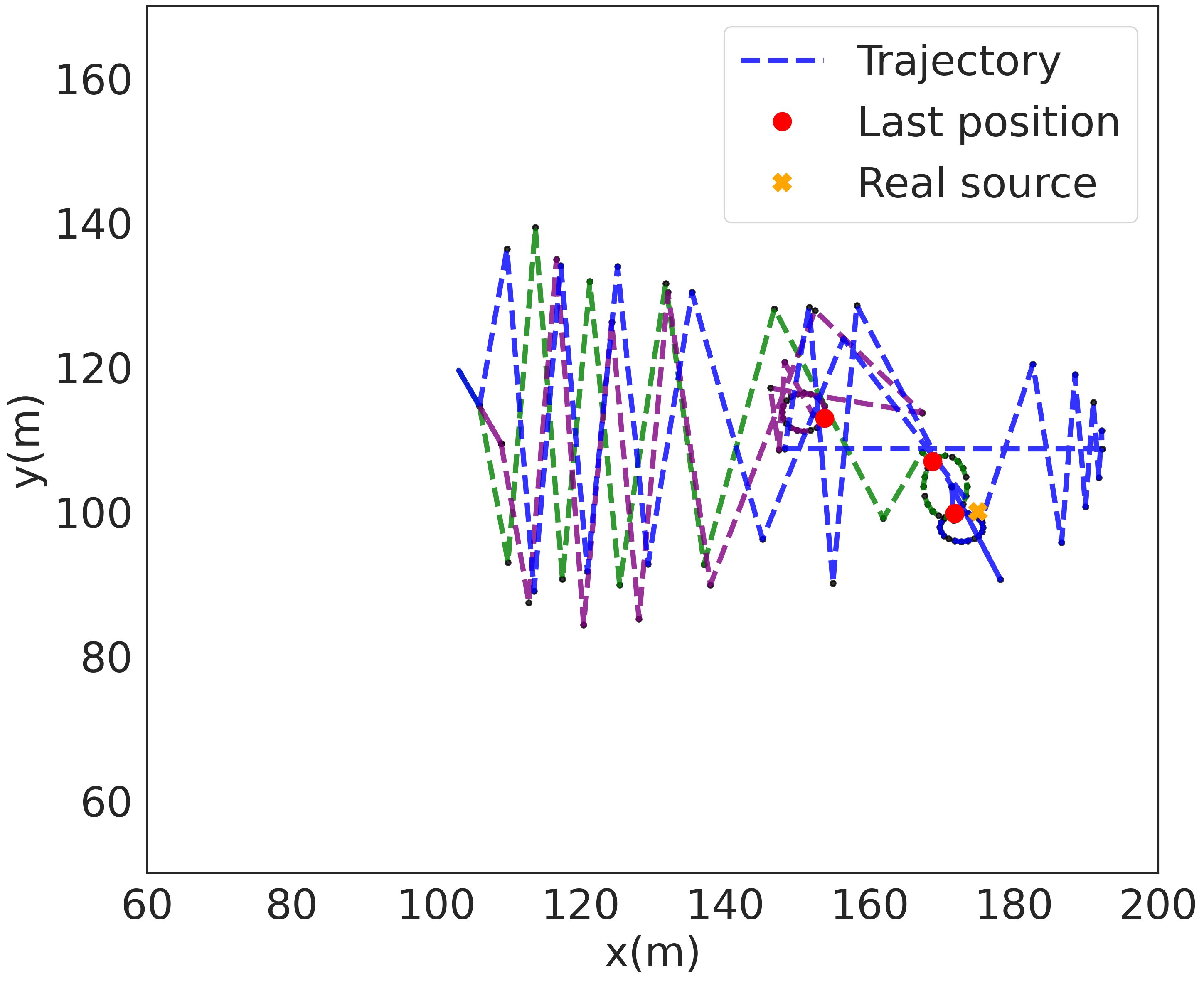}
        \caption{}
        \label{fig:t3}
    \end{subfigure}\\
       \begin{subfigure}{0.315\textwidth}
        \centering
        \includegraphics[width=\linewidth]{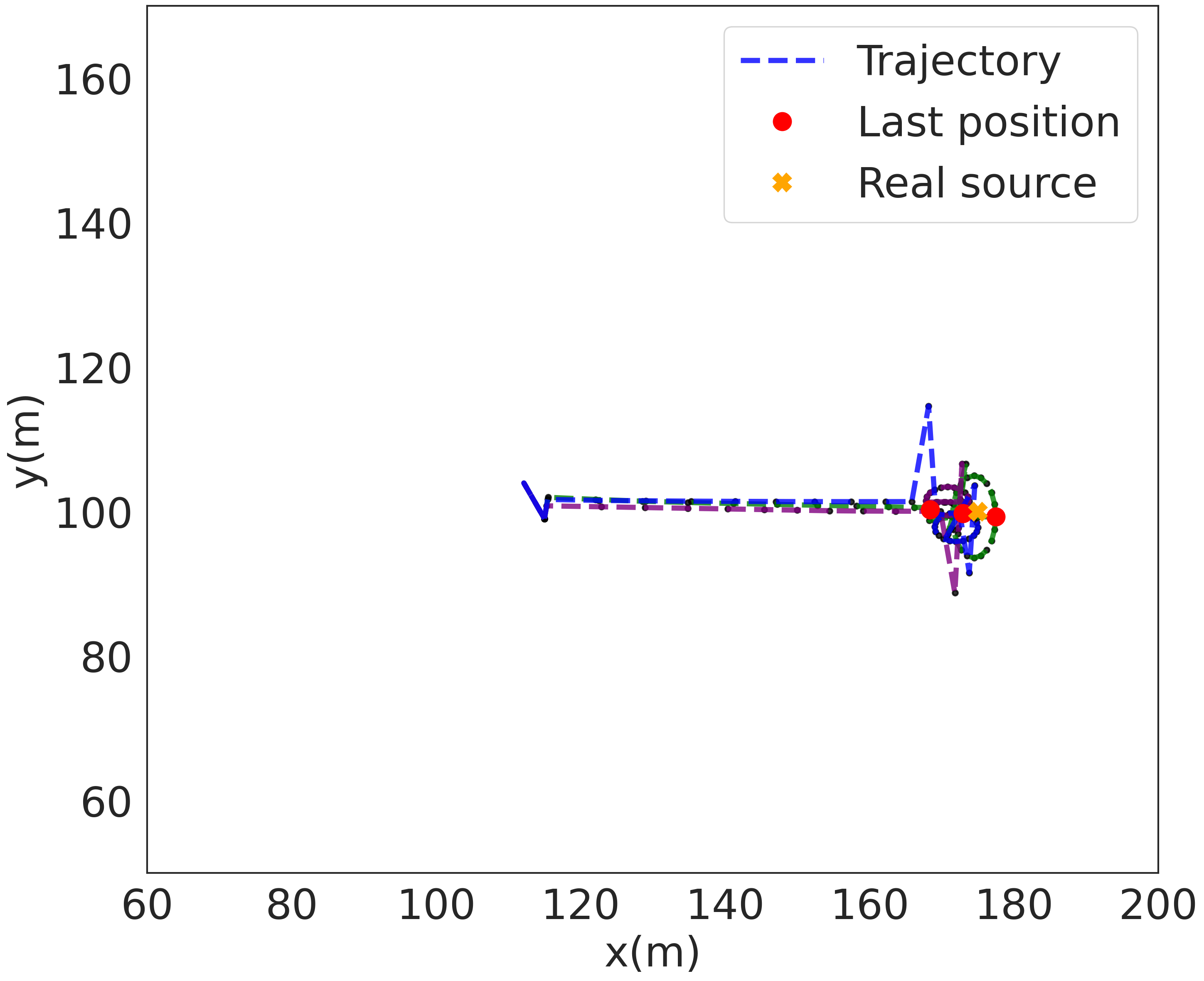}
        \caption{}
        \label{fig:t4}
    \end{subfigure}
    \hfill
    \begin{subfigure}{0.315\textwidth}
        \centering
        \includegraphics[width=\linewidth]{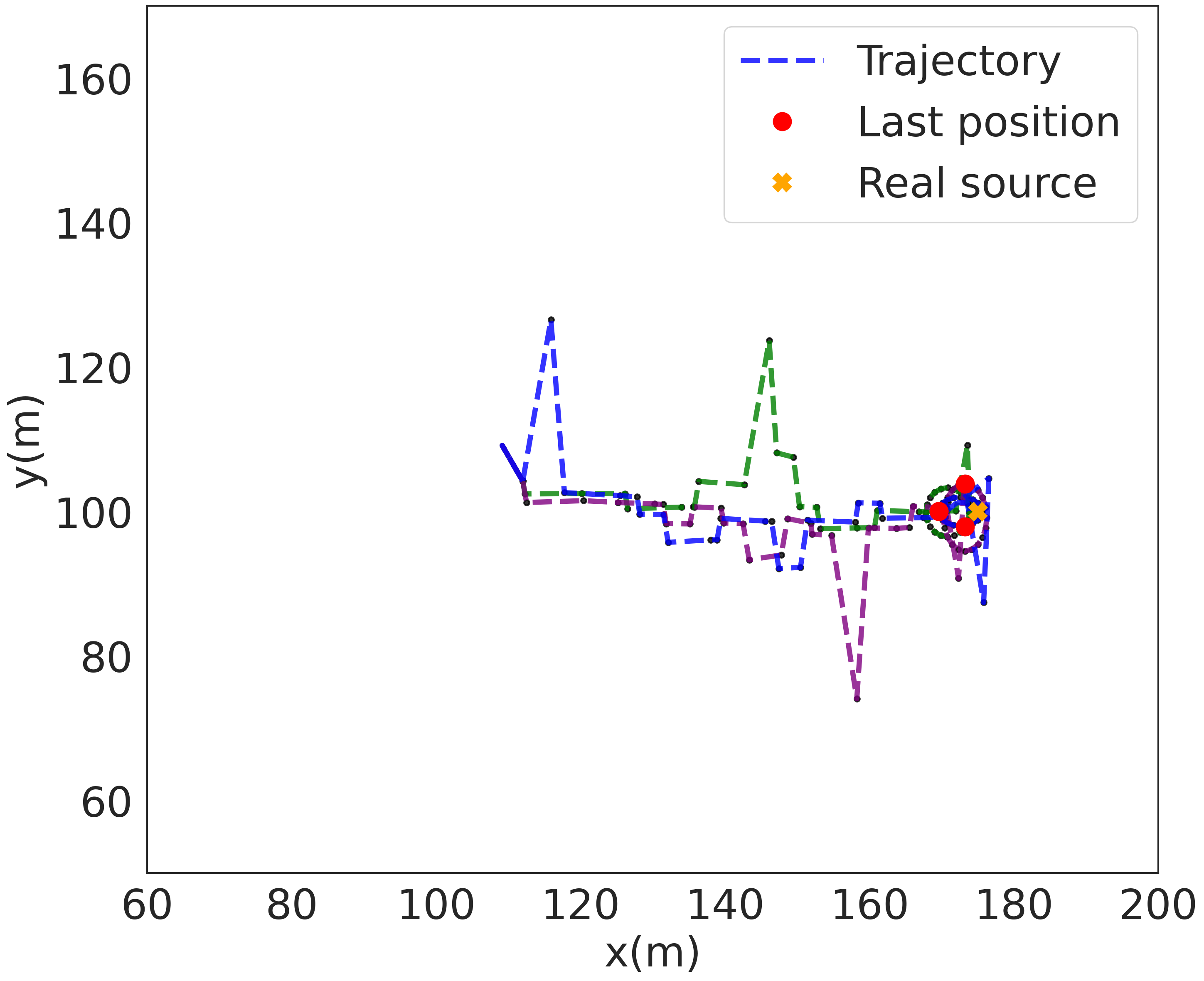}
        \caption{}
        \label{fig:t5}
    \end{subfigure}
    \hfill
    \begin{subfigure}{0.315\textwidth}
        \centering
        \includegraphics[width=\linewidth]{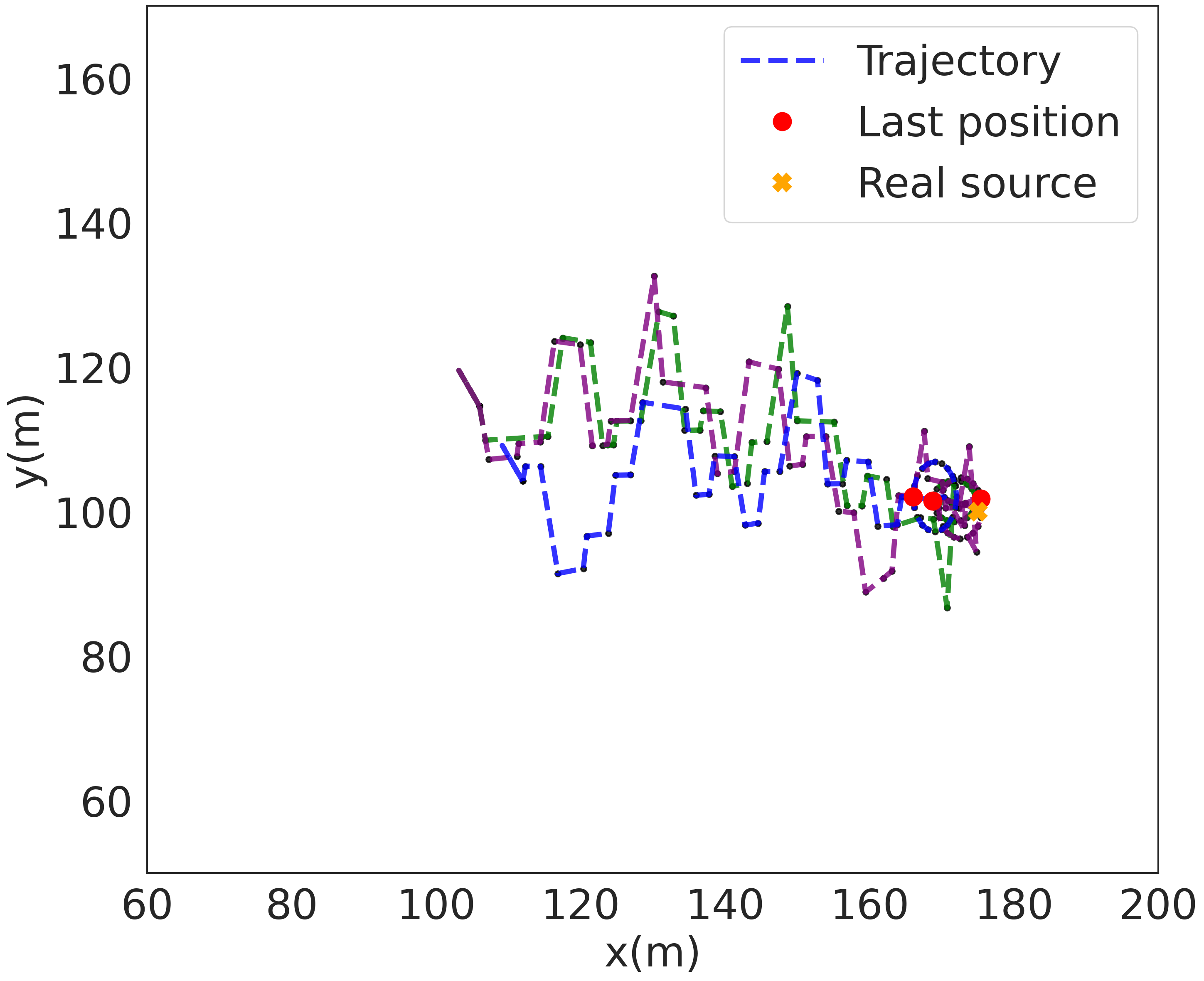}
        \caption{}
        \label{fig:t6}
    \end{subfigure}
    \caption{Representative trajectories of each scenario. (a) S1-F,  (b) S2-F,  (c) S3-F; (d) S1-H  (e) S2-H,  (f) S3-H.}
    \label{fig:trajectories}
\end{figure*}

The representative search trajectories shown in Figure~\ref{fig:trajectories} allow to compare Fast-Cognitive and Hybrid Fast-Cognitive across all scenarios, as well as to identify meaningful differences between their motion dynamics.

In Scenario 1 (Figures~\ref{fig:trajectories}a and \ref{fig:trajectories}d), the stable plume structure produces similar overall patterns for both methods, with trajectories remaining largely within the active plume region. Fast-Cognitive exhibits longer early-stage crossflow motions due to initial belief uncertainty, converging toward the source as uncertainty diminishes. The hybrid strategy shows a more directed path, dominated by source-directed motions with fewer exploratory deviations. Because the agent maintains consistent plume contact in stable conditions, detection-triggered exploitation produces efficient, direct trajectories.

In Scenario 2 (Figures~\ref{fig:trajectories}b and \ref{fig:trajectories}e), increased turbulence increases search uncertainty, leading to broader exploration in Fast-Cognitive trajectories. Crossflow movements become more pronounced as the agent searches within the predicted plume region. The hybrid strategy effectively balances exploitative source-directed movements upon detection with belief-informed exploration when contact is lost. This reactive switching reduces early-stage exploration while maintaining reliability, resulting in more goal-directed motion compared to the purely belief-driven Fast-Cognitive approach.

Scenario 3 (Figures~\ref{fig:trajectories}c and \ref{fig:trajectories}f) presents the greatest challenge due to high turbulence and plume intermittency. Fast-Cognitive shows extensive crossflow displacements driven by persistent belief uncertainty and broader inferred plume geometries. Exploitative behavior only emerges when the agent approaches the source, after substantial exploratory effort. The hybrid strategy shows markedly distinct dynamics where detection-triggered switching produces alternating patterns of source-directed motion during chemical contact and focused crossflow search when contact is lost. This reactive mechanism reduces the extensive lateral exploration characteristic of Fast-Cognitive, concentrating movement along the plume structure and achieving more direct source approach despite challenging conditions.

In general, H generates trajectories that are less exploratory and more directed towards the source, which translates into higher search efficiency. While this behaviour could suggest reduced reliability, the success rate shows the opposite, with H able to locate the source more frequently than the other methods.

\subsection{Field Experiments}

\subsubsection{Experimental Setup}

\begin{figure*}[tbp]
    \centering
    \begin{subfigure}{0.47\textwidth}
        \centering
        \includegraphics[width=\linewidth]{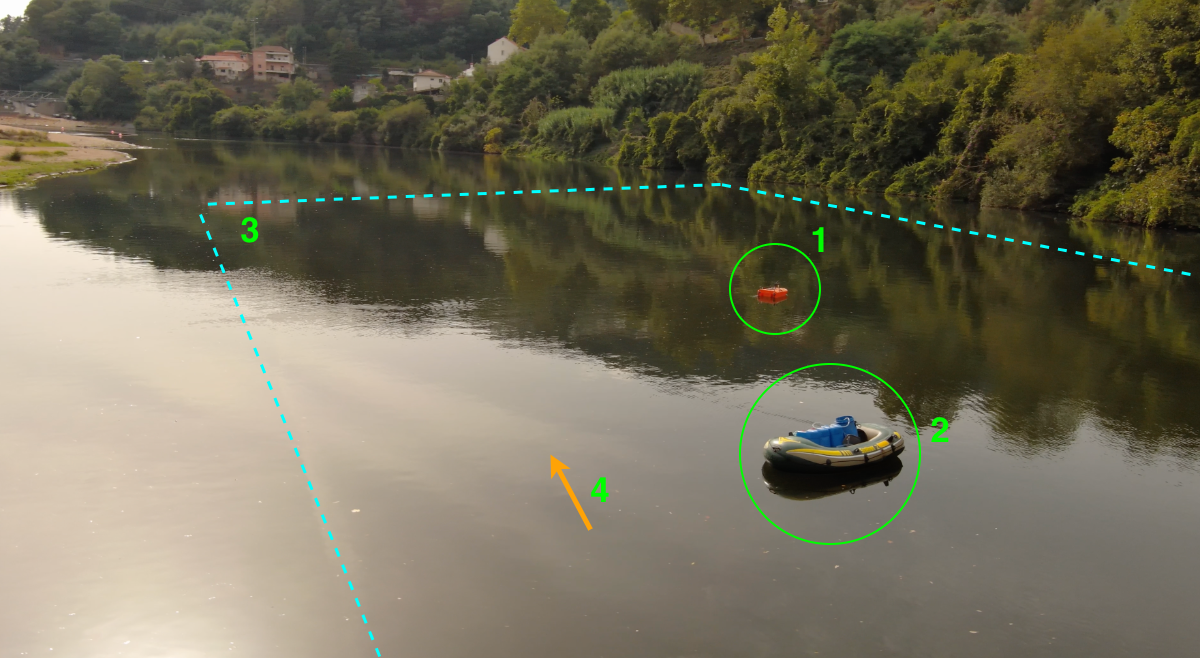}
        \caption{}
        \label{fig:f1}
    \end{subfigure}
    \begin{subfigure}{0.345\textwidth}
        \centering
        \includegraphics[width=\linewidth]{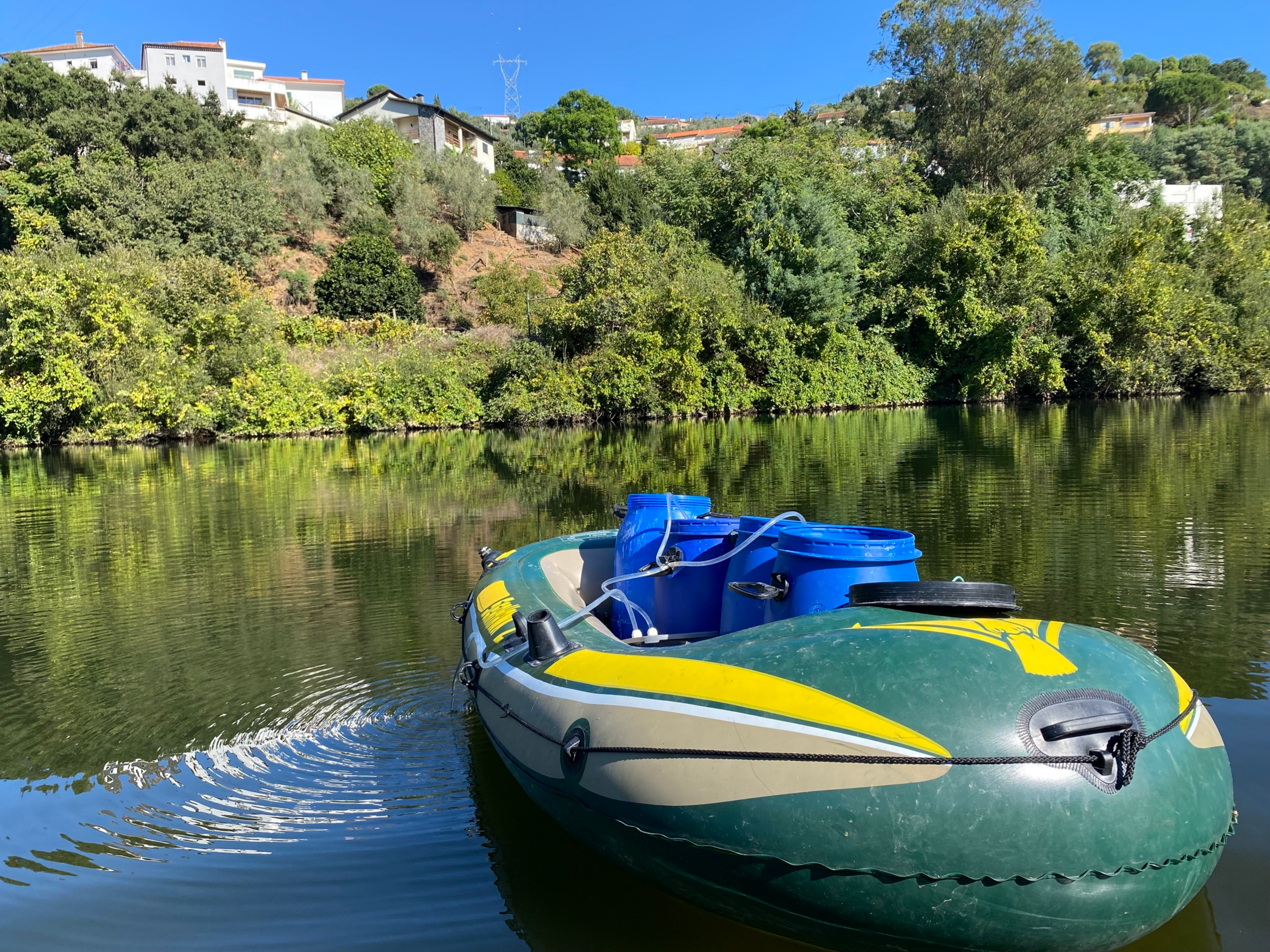}
        \caption{}
        \label{fig:f2}
    \end{subfigure}
    \caption{(a): OSL experiment in Mondego River with the ASV starting at a downflow location from the source. 1-ASV,  2-odour source,  3-testing section,  4-flow direction; (b): Anchored boat containing reservoirs of saturated salted water a peristaltic pump delivering the solution into the river flow.}
    \label{fig:fieldenv}
\end{figure*}

The field validation was conducted in the Mondego River near the University of Coimbra, Portugal (Figure~\ref{fig:fieldenv}a),  with an ASV as the searching agent. The testing section has a length of approximately 100 m per 60 m of width and an average depth of 2 m. The measured conductivity value of clean water $c_{b}$ was $45~\mu S/cm$ (baseline), the water temperature was at $22^\circ C$ and the flow speed at approximately $1.5$~m/s. All subsequent conductivity measurements (odour concentration values) shown in this work represent variations between the absolute value $c_{t}$ and this baseline (i.e., $\Delta c = c_{t} - c_{b}$), where positive values indicate plume detection. A pollution event (odour plume) is simulated with a solution of salted water emitted at a constant rate of 1 l/min with a peristaltic pump from a stationary floating platform (Figure~\ref{fig:fieldenv}b). The solution is a mixture of 1 kg of salt for 4 l of water, with each reservoir having a total volume of 30 l, allowing 30 min of experimental time. The salted water produces a realistic non-pollutant odour plume easily distinguished from clean water by measuring variations of the water condutivity relative to its baseline. 

\begin{figure*}[tbp]
    \centering
    \begin{subfigure}{0.4\textwidth}
        \centering
        \includegraphics[width=\linewidth]{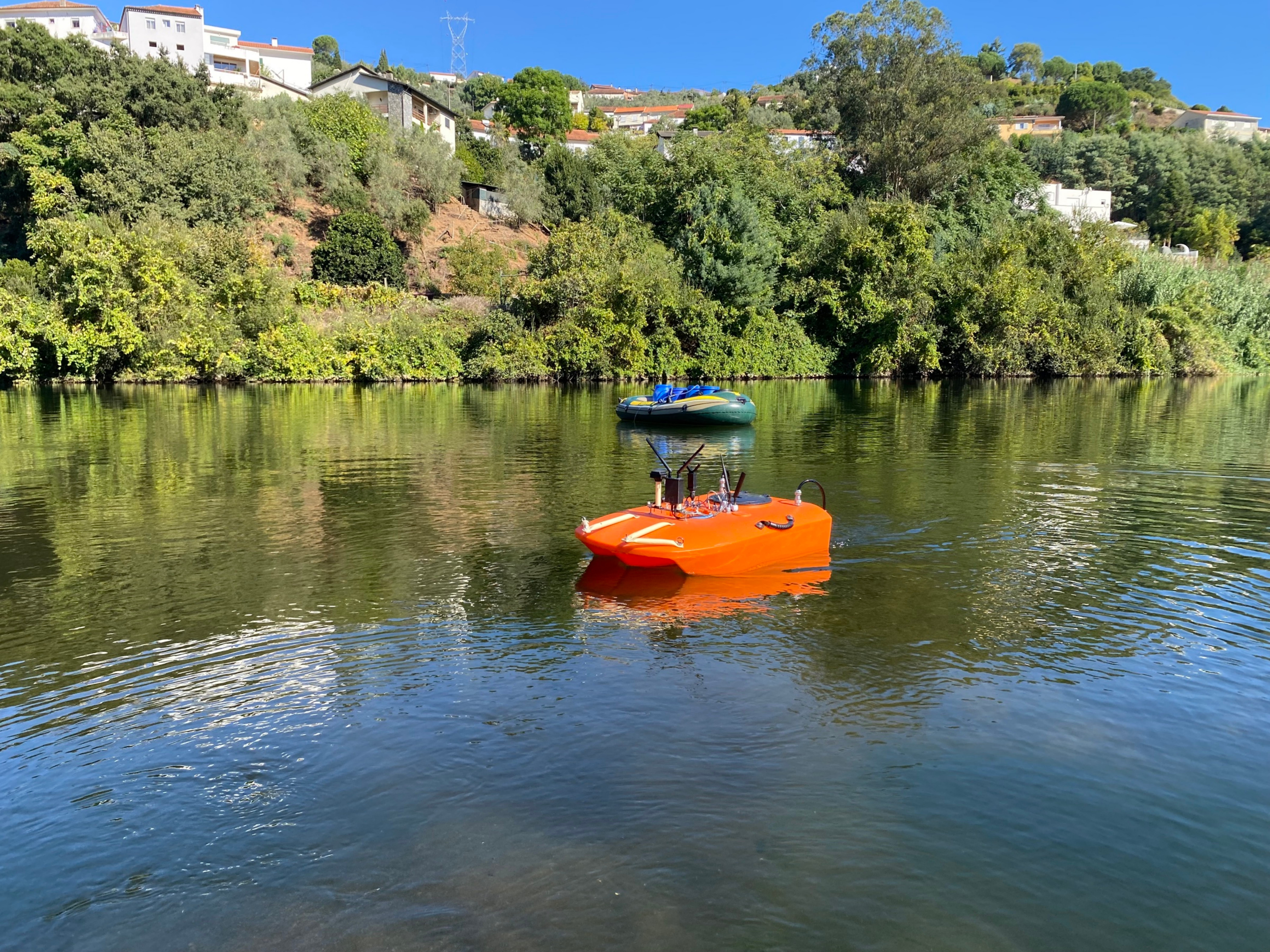}
        \caption{}
        \label{fig:f3}
    \end{subfigure}
    \begin{subfigure}{0.4\textwidth}
        \centering
        \includegraphics[width=\linewidth]{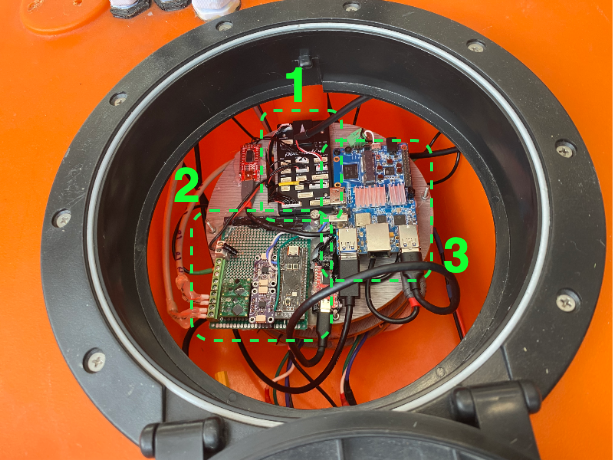}
        \caption{}
        \label{fig:f4}
    \end{subfigure}\\
    \centering
    \begin{subfigure}{0.4\textwidth}
        \centering
        \includegraphics[width=\linewidth]{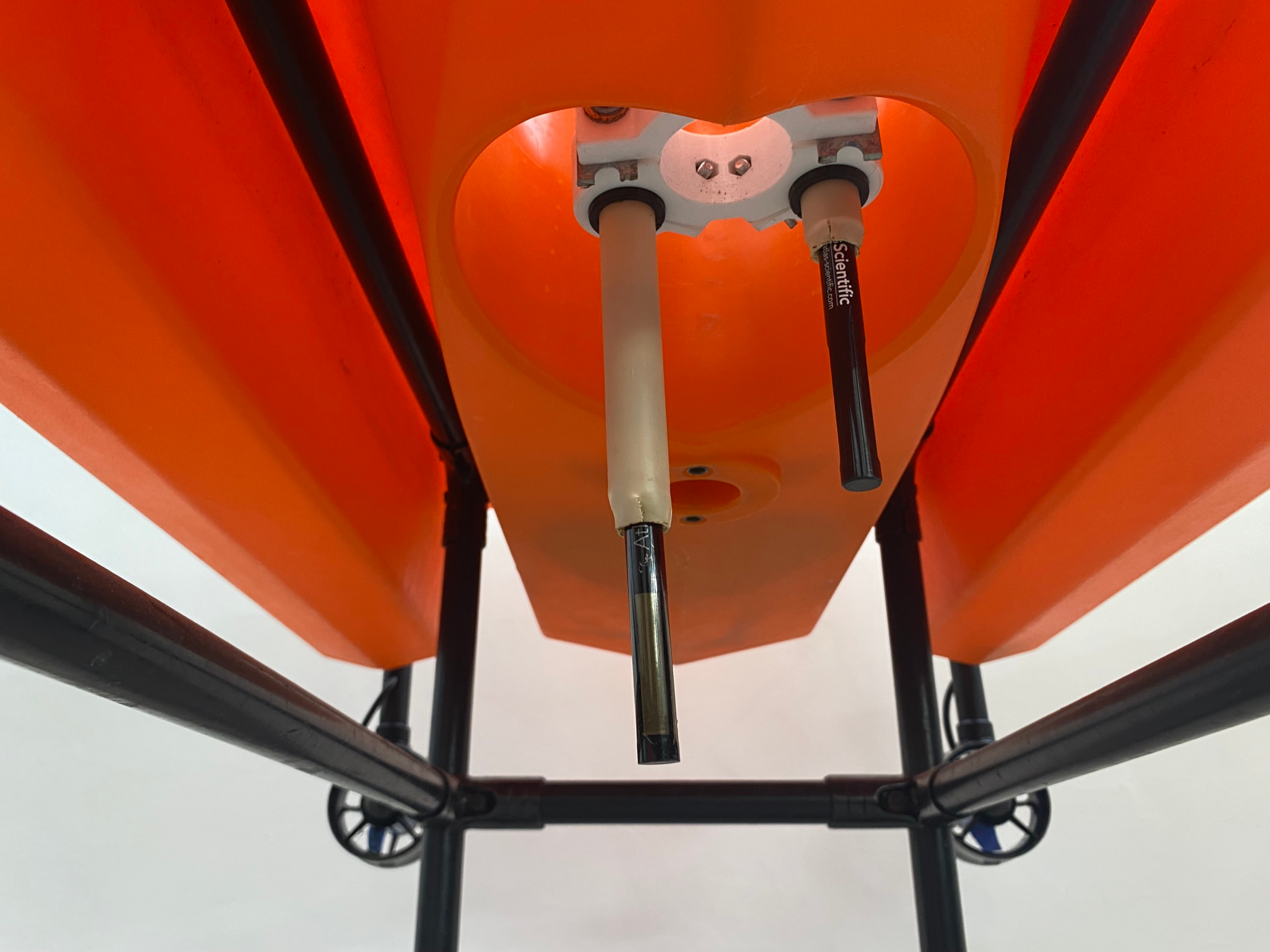}
        \caption{}
        \label{fig:f5}
    \end{subfigure}
    \begin{subfigure}{0.4\textwidth}
        \centering
        \includegraphics[width=0.65\linewidth]{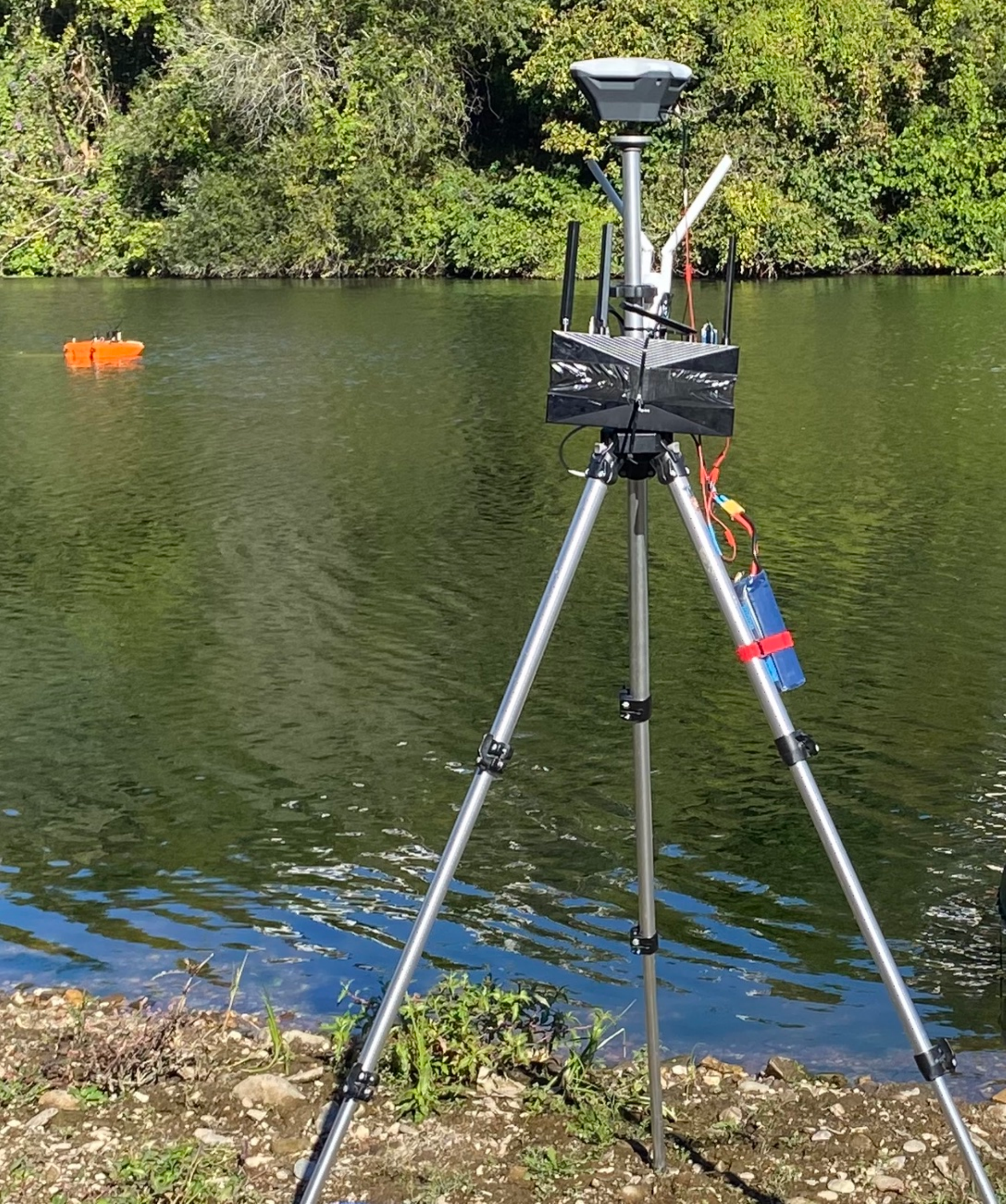}
        \caption{}
        \label{fig:f6}
    \end{subfigure}
    \caption{Elements of the field experiments. (a): ASV navigating in the river; (b): electronics on-board the ASV, 1-Pixhawk 4 autopilot, 2-Sensor acquisition circuit board, 3-Orange pi 3 SBC; (c) conductivity sensors installed at different depths and bottom view of the boat in a supporting structure; (d) base station assuring Wi-Fi communications between the ASV and an external laptop monitoring the missions, while also sending RTK localisation corrections directly to the searching agent.}
    \label{fig:field_elements}
\end{figure*}

The ASV performing the OSL mission is composed of a fibreglass hull with an overall dimension of 0.8 x 0.6 m (Figure~\ref{fig:field_elements}a). It is equipped with two Bluerobotics T200 thrusters attached at the back allowing for differential motion navigation and a maximum speed of 2.0 m/s. Localisation is achieved with a Global Navigation Satellite System (GNSS) module, configured in real-time kinematic positioning mode. The platform's motion is controlled by a Pixhawk 4 autopilot maintaining a moving speed of 0.5 m/s between goal positions. The high-level mission is provided by an Orange Pi 3 single board computer (SBC) running a Robot Operating System (ROS) framework. The algorithms were coded with C++ and Python programming languages, with the most computationally demanding functions optimised with Numba JIT compiler. 

A 4S LiPo battery with an 18 Ah capacity assures full operational autonomy between 6 and 8 hours. Two conductivity sensors from Atlas Scientific with a measurement range between 0.07 and 50,000 $\mu S/cm$ are installed at the centre of the hull at different depths (0.2 and 0.5 m) (Figure~\ref{fig:field_elements}c), allowing the measurement of small variations originated from the salted plume.  Only the sensor positioned at the highest depth is used during these searching missions. A custom-made circuit board with an STM32F4 microcontroller enables the low-level interface with the sensor and sends the information to the SBC through USB communication at a rate of 2 Hz. The Pixhawk 4, SBC and sensor circuit board can be observed in Figure~\ref{fig:field_elements}b. The platform also uses a Bluerobotics Ping Sonar to measure the depth of the river basin. A small communications base station (Figure~\ref{fig:field_elements}d) is installed on a tripod at the margin of the river enabling localisation corrections through an RTK base, and Wi-Fi communication between the ASV and an external monitoring laptop with an outdoor Access Point.

Seven experiments were conducted with the agent starting approximately 40~m downstream from the source and outside the active plume region. A small circular exclusion region with a radius of 2 m was defined at the true source location to prevent the agent from colliding with the floating platform. If the trajectory between the agent and a goal position crosses this region, the agent circumnavigates it. When a goal position lies inside the exclusion region and the agent reaches its boundary, the source is considered found.

\subsubsection{Environmental Challenges}

The Mondego River presents substantial challenges that exceed controlled simulation conditions. Flow velocities varied between 1.0 and 2.0~m/s during experiments due to an upstream dam operations, causing rapid changes in plume structure and transport dynamics. The testing section features complex bathymetry including shallow regions and sandy banks that create localized flow disturbances. Natural turbulence produce concentration fluctuations substantially higher than simulated scenarios, with the salt-water plume exhibiting pronounced intermittency and fragmentation.

Unlike simulation environments, where plume models accurately predict concentration distributions, field conditions introduce persistent model-environment mismatch. The higher density of the saline solution causes vertical dispersion toward the riverbed, creating three-dimensional plume structures that are challenging to surface-based sensing. Furthermore, varying flow conditions during individual experiments can shift the plume structure mid-search, requiring strategies robust to dynamic environmental changes.

\subsubsection{Field Results}

The hybrid strategy's detection-triggered switching mechanism directly addresses these challenges. Rather than relying solely on belief accuracy, which degrades under model mismatch, reactive component ensures source-directed and belief-based motion upon chemical contact regardless of belief state. This provides increased robustness to the estimation errors inherent in complex field environments, and guides the agent towards higher concentration regions more rapidly that consequently improves the estimation process.

\begin{figure*}[tbp]
    \centering
    \begin{subfigure}{0.4\textwidth}
        \centering
        \includegraphics[width=\linewidth]{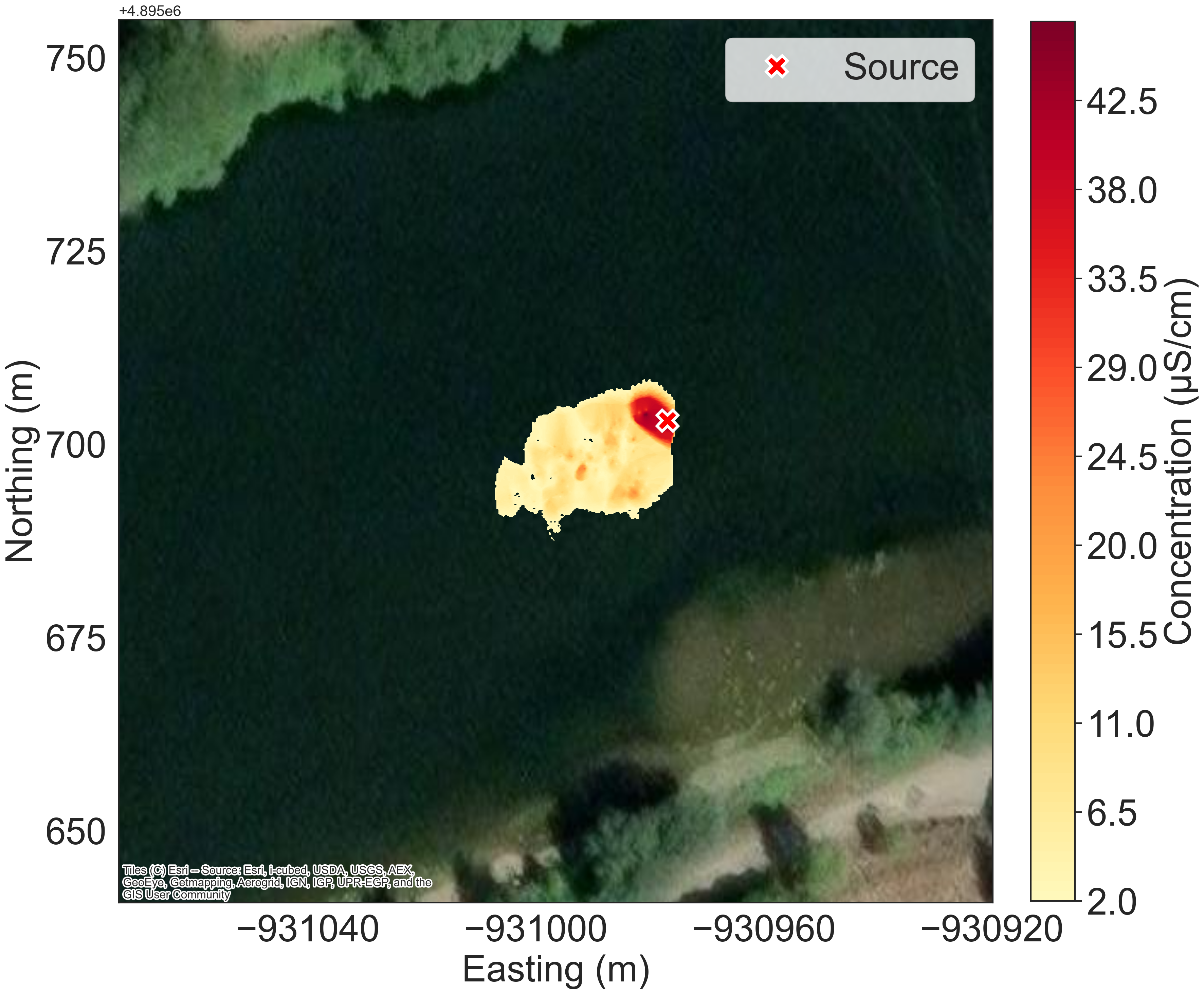}
        \caption{}
        \label{fig:f111}
    \end{subfigure}
    \begin{subfigure}{0.4\textwidth}
        \centering
        \includegraphics[width=\linewidth]{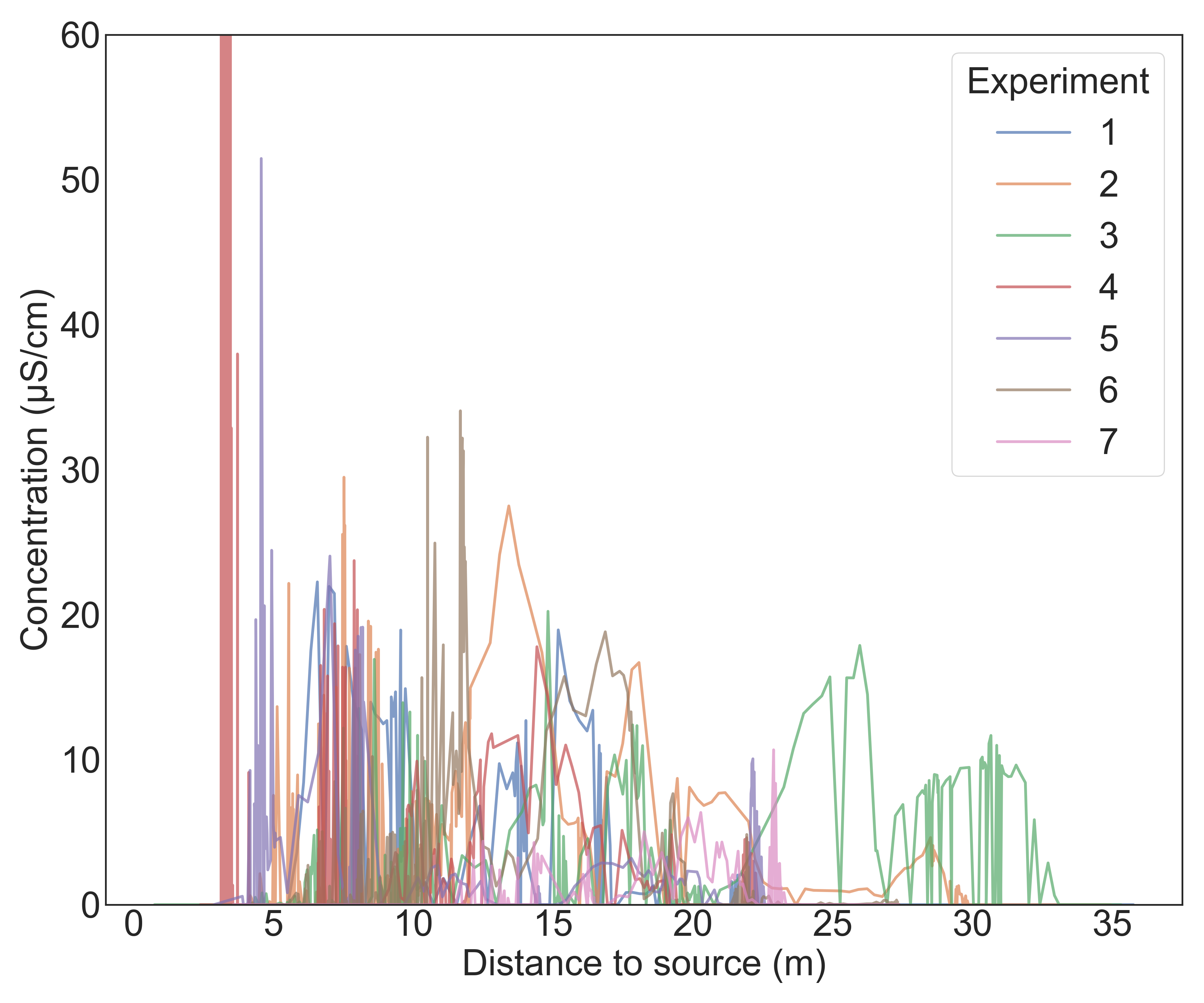}
        \caption{}
        \label{fig:f222}
    \end{subfigure}
    \caption{(a): Spatial interpolation map of all measured concentration variations from baseline during field experiments; (b): Measured concentration variations from baseline relative to the distance to source during each field experiment, acquired with a sampling rate of 2 Hz.}
    \label{fig:field_conc_map}
\end{figure*}

Spatial interpolation of measured concentrations (conductivity variations from the baseline) using the Inverse Distance Weighting method~\cite{lu2008adaptive} reveals substantial plume complexity (Figure~\ref{fig:field_conc_map}a). The concentration field exhibits characteristics broadly consistent with Gaussian dispersion, with decreasing concentrations with increasing downflow distance from source and lateral displacement from centerline. Although it also shows pronounced distortion and localized high-concentration clusters indicating intermittency driven by natural flow instabilities.

Figure~\ref{fig:field_conc_map}b shows the measured concentrations relative to distance from the source across all experiments. The general trend confirms the expected decay pattern, with concentrations reaching higher than 50--60~$\mu$S/cm above the baseline near the source, and decreasing substantially for distances beyond 20~m. However, the considerable variability between experiments and the irregular fluctuations within individual trajectories highlight the intermittent nature of field plumes. In some experiments (e.g., experiment 3) it can be observed detections at distances exceeding 30~m, while others lost contact closer to the source, reflecting the dynamic plume structure created by varying flow conditions.
   
Six of seven experiments successfully located the source, yielding an 86\% success rate that validates the hybrid strategy under real-world conditions. Table~\ref{tab:exp_field_metrics} summarizes performance metrics across all trials.

\begin{table*}[tbp]
\centering
\caption{Summary of field experiment performance metrics}
\label{tab:exp_field_metrics}
\begin{tabular}{lccc}
\toprule
\textbf{Metric} & \textbf{Min} & \textbf{Mean} & \textbf{Max} \\
\midrule
Dist. to source, first detection (m) & 21.35 & 24.21 & 30.59 \\
Total travel distance (m) & 36.07 & 76.33 & 117.17 \\
Distance ratio & 1.69 & 3.12 & 4.34 \\
\midrule
Total comput. time per experiment (s) & 1.92 & 3.42 & 5.40 \\
Total movement time (s) & 72.13 & 152.65 & 234.34 \\
Total search time per experiment (s) & 74.06 & 156.07 & 239.75 \\
\midrule
Localisation error (m) & 0.35 & 3.17 & 8.49 \\
Localisation uncertainty (m) & 0.91 & 7.34 & 23.44 \\
\midrule
Total decisions (n) & 12 & 16 & 25 \\
Behavior switches (n) & 8 & 9.86 & 14 \\
Behavior continuity (n) & 3 & 5.14 & 13 \\
Crossflow decisions (n) & 5 & 7.86 & 12 \\
Source directed decisions (n) & 5 & 7.57 & 12 \\
\bottomrule
\end{tabular}
\end{table*}

The average localization error of 3.17~m shows accurate source identification despite environmental complexity, with the minimum error approaching zero in experiments where the agent directly encountered the source. The maximum error of 8.49~m occurred in the single failed experiment, where belief divergence led to incorrect source estimation.

The distance ratio quantifies search efficiency by comparing traveled distance to straight-line distance from first detection to source. The average ratio of 3.12, with best performance reaching 1.69, represents substantial improvement over Fast-Cognitive field experiments in~\cite{magalhaes2025fastcog}, which exhibited ratios between 6.0 and 8.0 under similar conditions. This improvement confirms that reactive belief-based decisions reduces over-exploration in real-world environments, consistent with simulation findings.

Behavioral switching patterns confirm the hybrid mechanism operates as designed under field conditions. On average, experiments required 16 movement decisions, with behavior 
switches (9.86) occurring more frequently than behavior continuations (5.14). This switching frequency reflects the intermittent nature of field plumes, with the agent alternating between source-directed exploitation upon chemical contact and crossflow 
exploration when contact is lost. The near-equal distribution between crossflow (7.86) and source-directed (7.57) decisions indicates balanced exploitation-exploration, with neither mode dominating despite the reactive switching mechanism.

Computational performance confirms real-time feasibility on embedded hardware. Total computation time averaged 3.42~s per experiment, representing only 2.2\% of total search time (156.07~s). This minimal computational overhead reveals that the hybrid strategy maintains the efficiency advantages of Fast-Cognitive while adding reactive capabilities.

\subsubsection{Analysis of Trajectories}
\begin{figure*}[tbp]
    \begin{subfigure}{0.32\textwidth}
        \centering
        \includegraphics[width=\linewidth]{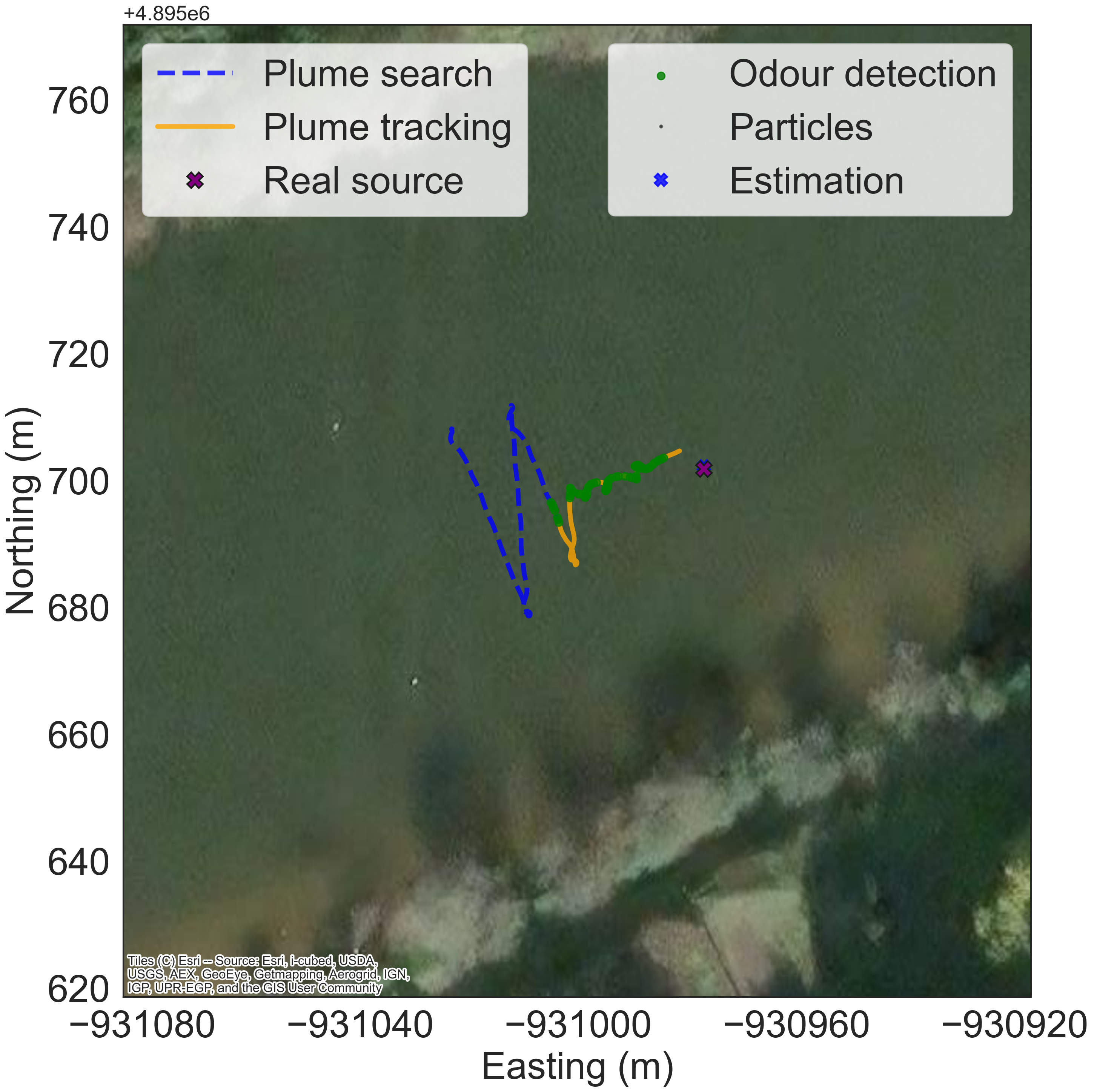}
        \caption{}
        \label{fig:f7}
    \end{subfigure}
    \begin{subfigure}{0.32\textwidth}
        \centering
        \includegraphics[width=\linewidth]{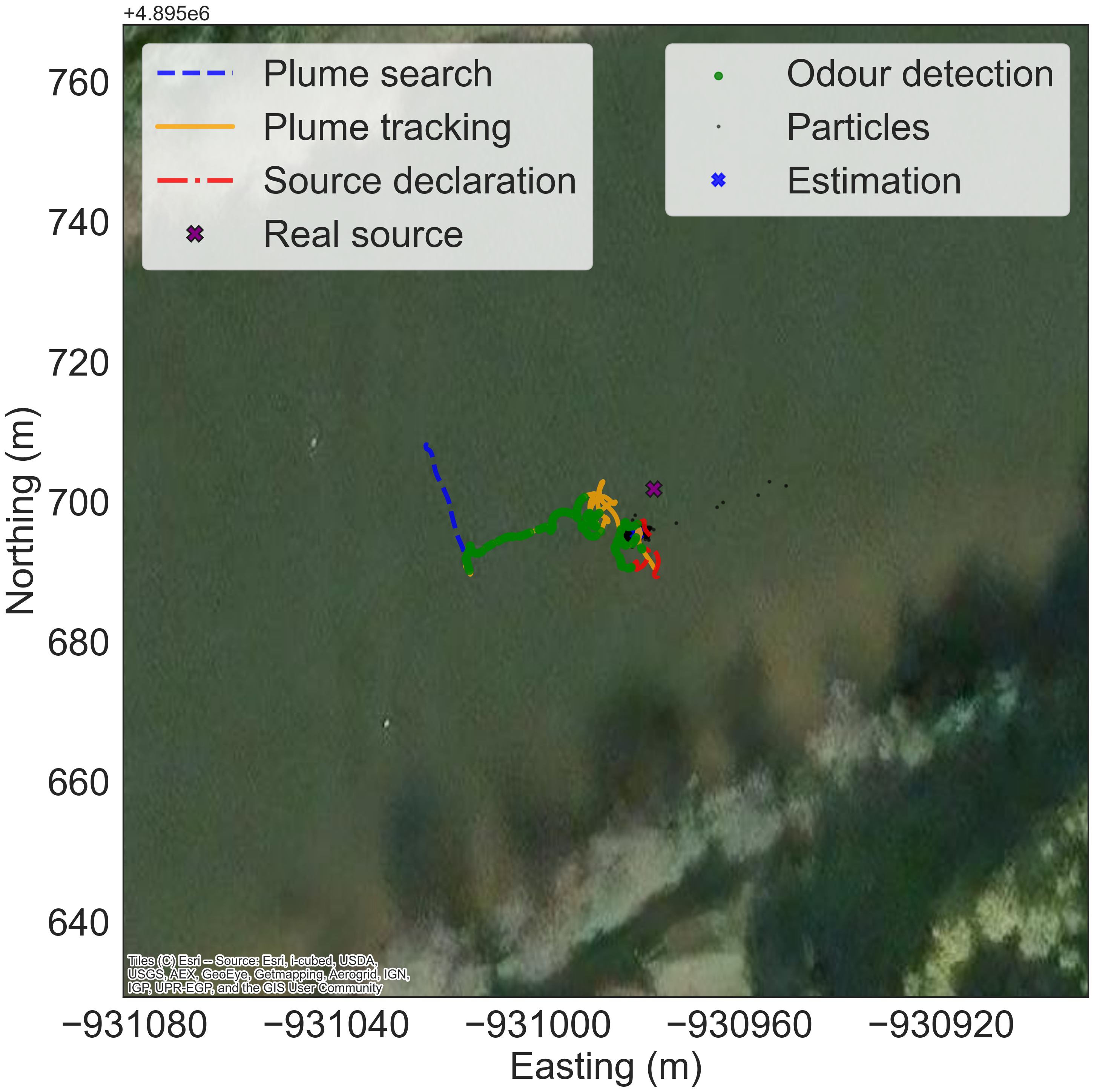}
        \caption{}
        \label{fig:f8}
    \end{subfigure}
    \begin{subfigure}{0.32\textwidth}
        \centering
        \includegraphics[width=\linewidth]{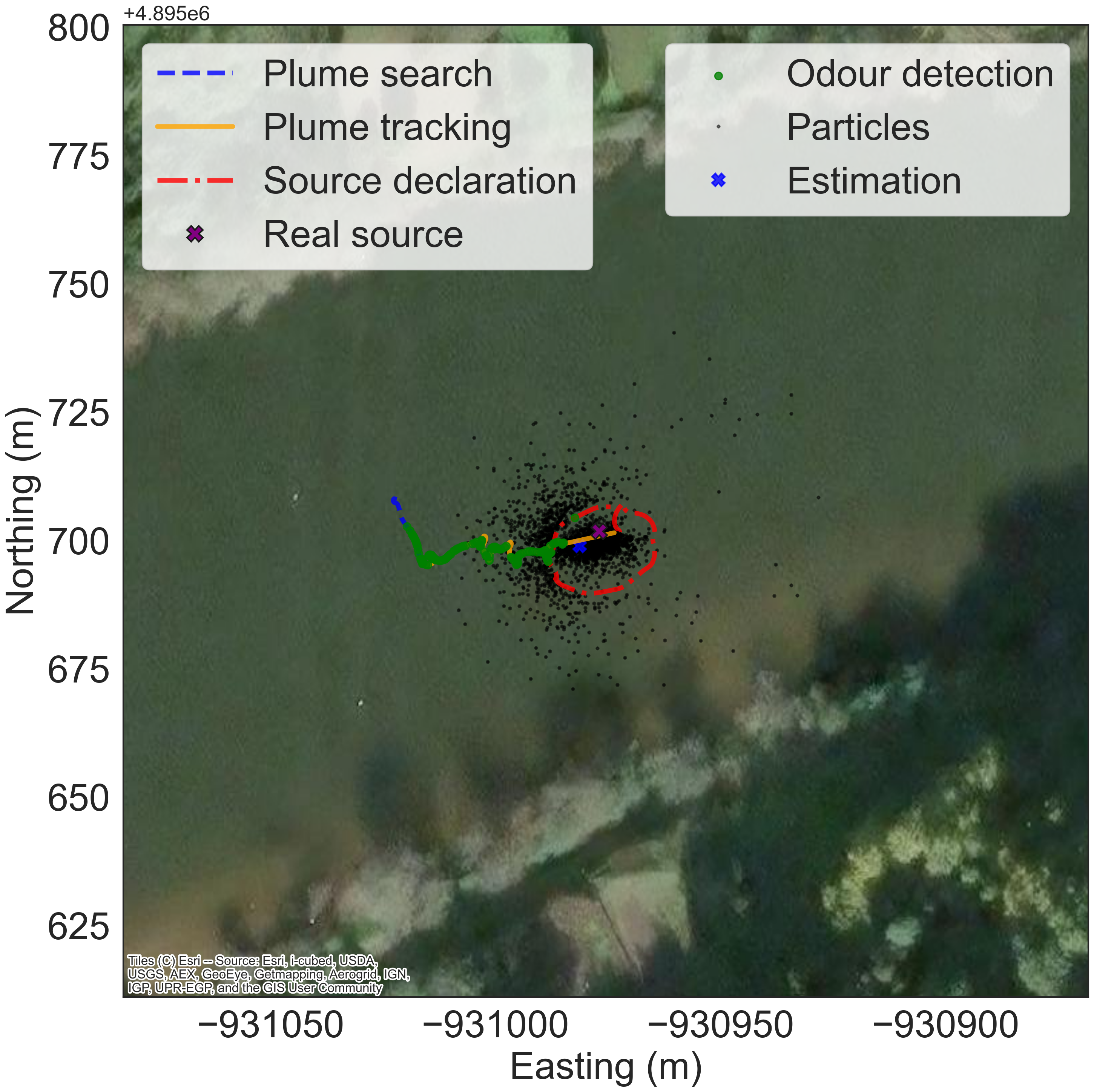}
        \caption{}
        \label{fig:f88}
    \end{subfigure}\\
    \begin{subfigure}{0.32\textwidth}
        \centering
        \includegraphics[width=\linewidth]{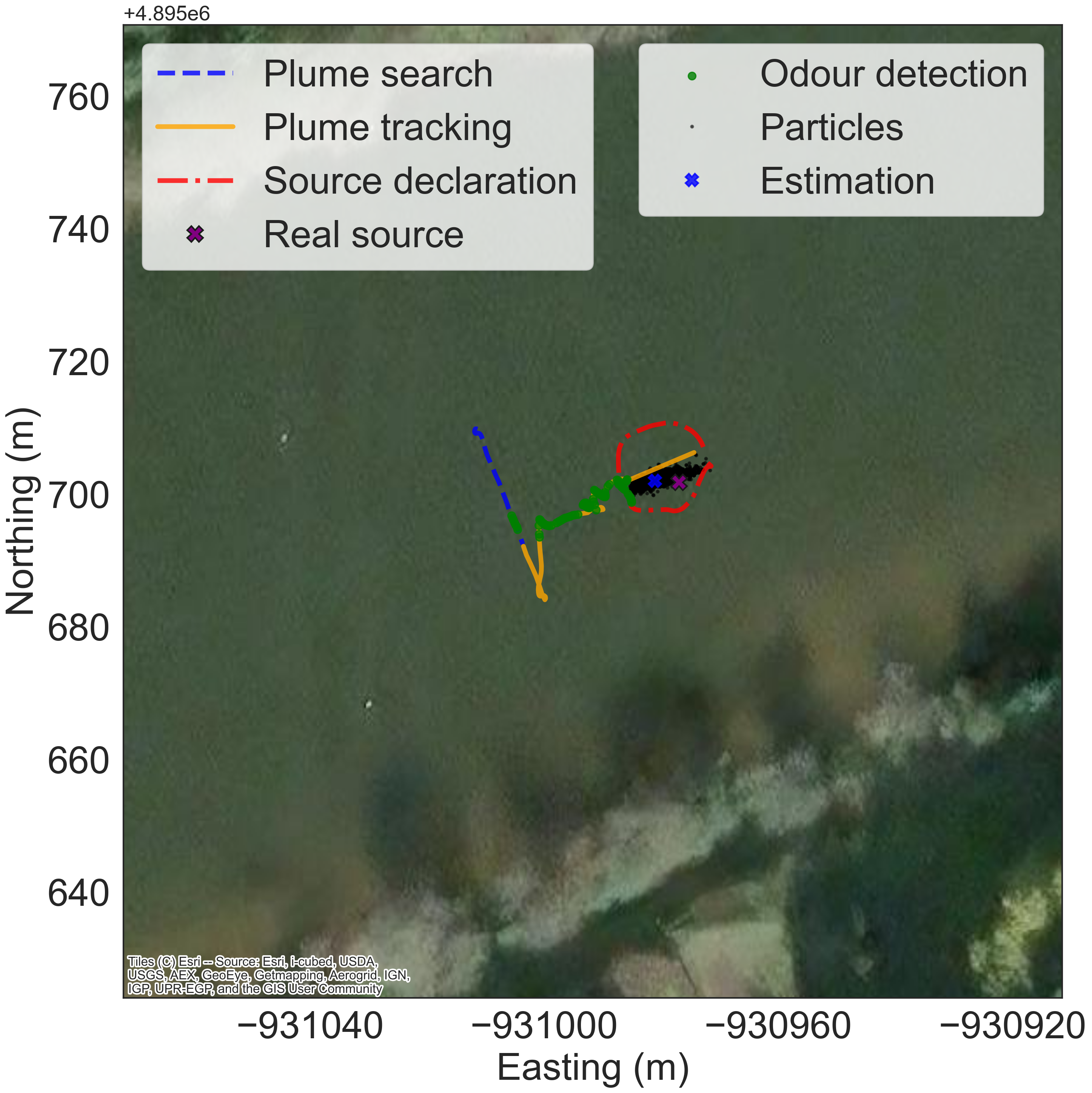}
        \caption{}
        \label{fig:f9}
    \end{subfigure}
    \begin{subfigure}{0.32\textwidth}
        \centering
        \includegraphics[width=\linewidth]{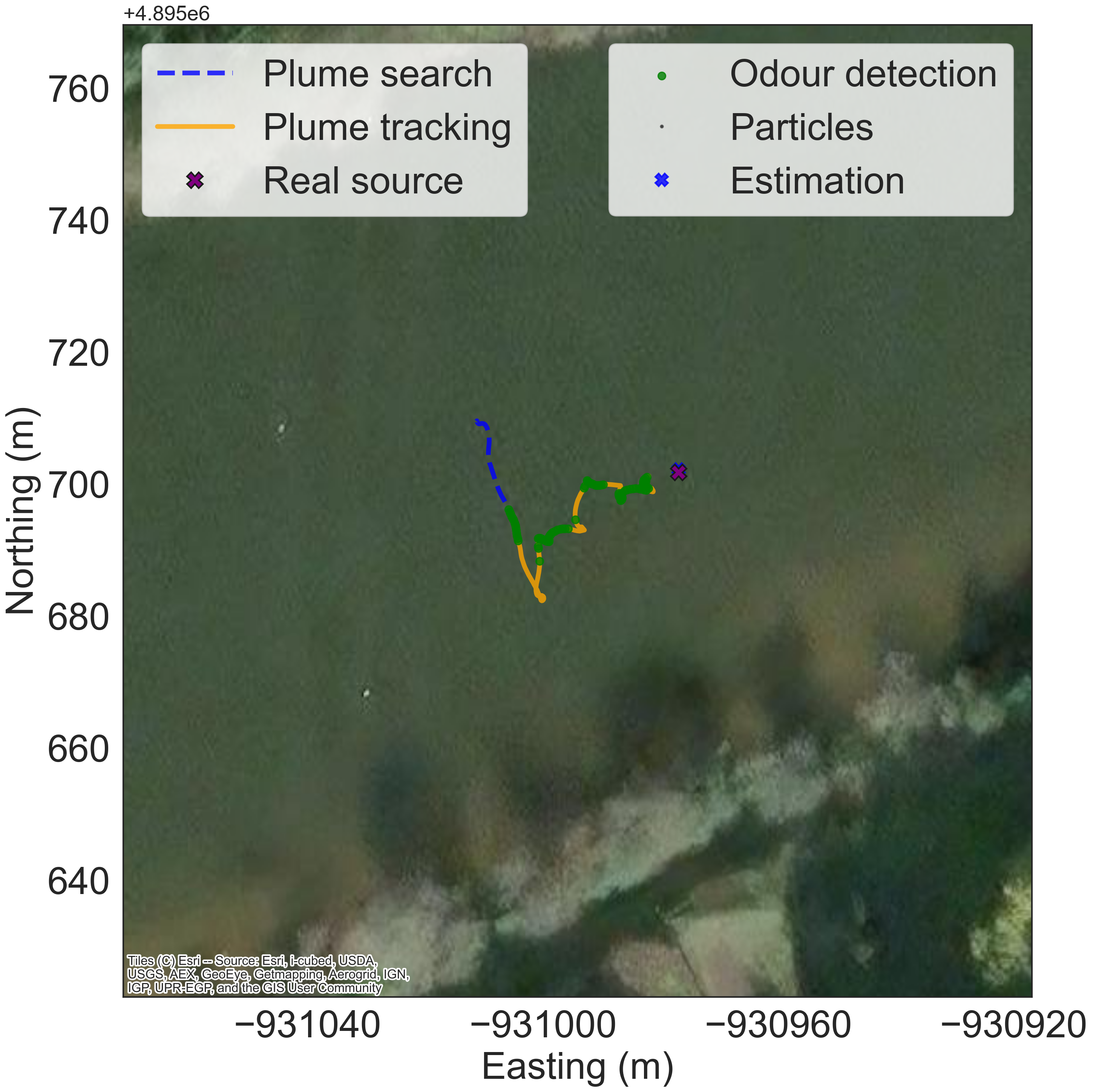}
        \caption{}
        \label{fig:f10}
    \end{subfigure}
    \begin{subfigure}{0.32\textwidth}
        \centering
        \includegraphics[width=\linewidth]{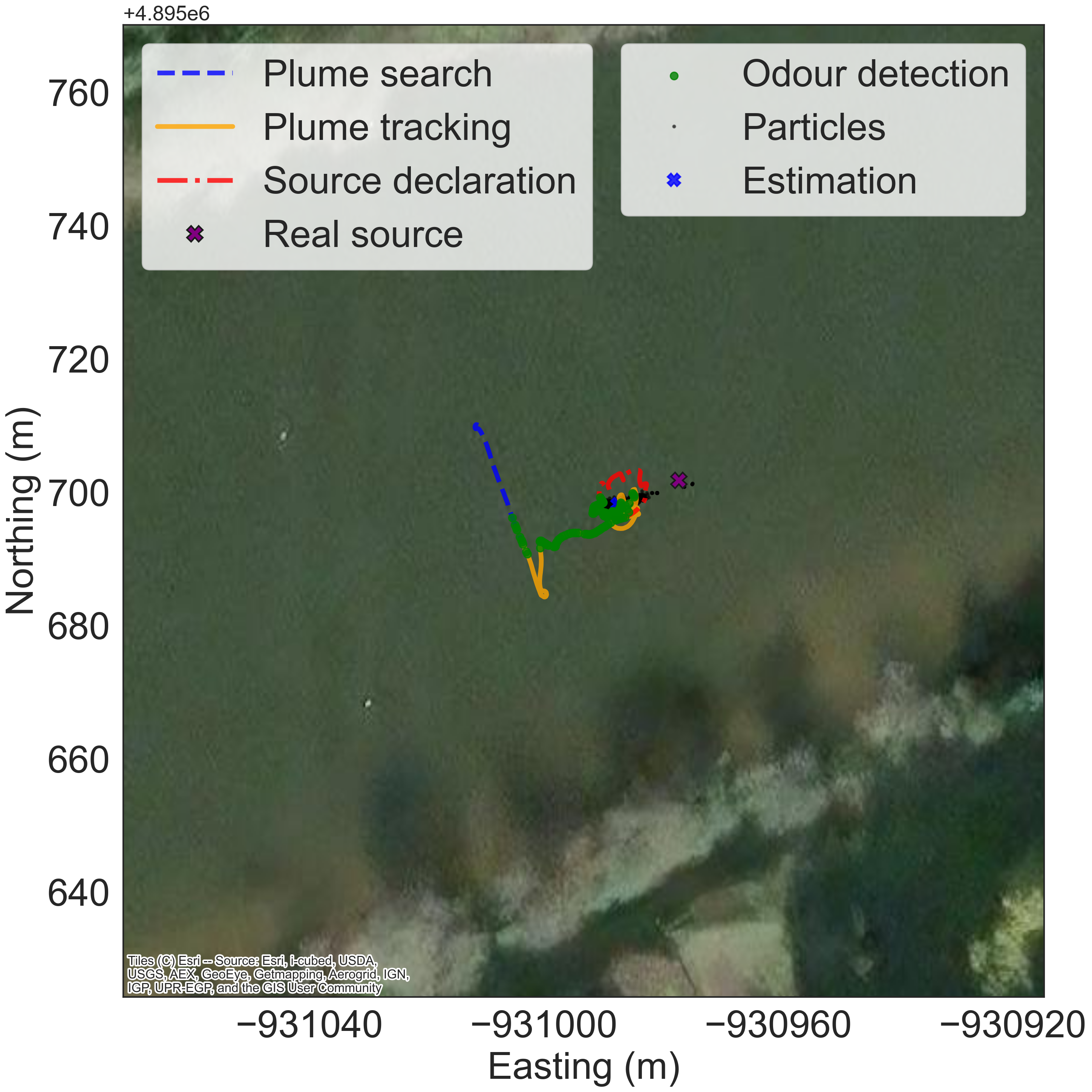}
        \caption{}
        \label{fig:f100}
    \end{subfigure}
    \caption{Example trajectories of the proposed algorithm in a field scenario.}
    \label{fig:field_trajectories}
\end{figure*}

Figure~\ref{fig:field_trajectories} presents search trajectories of six field experiments, illustrating the hybrid strategy's behavior under varying conditions.

The first experiment (Figure~\ref{fig:field_trajectories}a) reveals optimal hybrid performance, producing a highly directed trajectory compared to traditional cognitive approaches. After initial crossflow movements to locate the plume, the agent alternates between crossflow and source-directed behaviors based on chemical contact. This reactive switching guides the agent away from low-intensity measurements distant from the source while maintaining proximity to the active plume region. Unlike Fast-Cognitive and Infotaxis, which maintain exploratory tendencies until uncertainty decreases substantially, the hybrid approach prioritizes direct source localization over belief refinement. The resulting trajectory achieved exact source localization without requiring declaration behavior, as the agent moved directly to the true source position.

The second experiment (Figure~\ref{fig:field_trajectories}b) represents the single failed case. The agent detected odorant near the plume centerline and executed numerous source-directed motions throughout the search. However, upon approaching the source, plume contact was lost. Turbulence-induced transitions between crossflow and source-directed motions directed the agent away from the source, causing premature belief convergence lateral to the active region. Consequently, the declaration behavior occurred far from the true source, and measurements collected during this phase failed to correct the belief. This failure illustrates a limitation of detection-triggered switching under highly intermittent conditions where frequent behavior changes can become counterproductive.

In the third experiment (Figure~\ref{fig:field_trajectories}c), the agent detected initial odorant at the active region's periphery, initiating localization with an offset from the plume 
centerline. The reactive mechanism enabled deviation correction without extensive exploration, where multiple crossflow behaviors produced upon losing contact corrected motion direction and maintained the agent within the active region. This represents an advantage over bio-inspired approaches such as Silkworm-Moth, where agents detecting odorant far from the centerline may be directed away from the source through upflow surge motions. The hybrid approach maintains bio-inspired reactivity while performing 
informed motions toward the estimated source. The agent reached the estimated position with low uncertainty and performed declaration behavior successfully around the true source, validating correct localization despite not crossing the exact source location.

The fourth experiment (Figure~\ref{fig:field_trajectories}d) initiated tracking closer to the plume centerline, enabling efficient movement toward the source due to frequent odour 
readings. Multiple crossflow behaviors corrected the direction determined from the estimated source location. Upon reaching the expected position, the agent successfully performed declaration behavior around the true source, with belief particles converged 
at this location. This trajectory exemplifies efficient hybrid operation when environmental conditions lead to consistent chemical contact.

The fifth experiment (Figure~\ref{fig:field_trajectories}e) involved fewer odorant contacts, reflected in longer crossflow motions followed by shorter source-directed decisions. The belief converged gradually toward the true source location, guiding the agent to successful localization without requiring declaration behavior. This experiment achieved one of the highest search efficiency among successful trials, revealing that the hybrid mechanism can produce direct source approach even with intermittent detections.

The sixth experiment (Figure~\ref{fig:field_trajectories}f) experienced premature belief convergence due to numerous source-directed behaviors that shifted the estimate downstream of the true source. Consequently, the declaration trajectory occurred without surrounding the source location. However, measurements collected during this final phase enabled belief convergence closer to the true source, allowing this experiment to be considered as successful despite the sub-optimal declaration placement.

These trajectories confirm that the hybrid decision-making performs effectively under real-world conditions, producing search patterns consistent with simulation results. 
The alternating exploitation-exploration dynamics reduce the extensive lateral movements characteristic of purely cognitive approaches, while the belief-driven parameterization maintains adaptive behavior across varying environmental conditions.

\subsubsection{Comparison with Simulation Results}

Field performance validates simulation predictions under challenging conditions. The 86\% field success rate (6 of 7 experiments) aligns with the 93\% achieved in S3 simulations, with the difference attributable to the limited number of experiments, additional field complexities including three-dimensional plume dispersion, dynamic flow variations from upstream dam operations, and model-environment mismatch inherent in natural environments.

The field distance ratio of 3.12 closely matches the S3 simulation value of 3.19 for the hybrid strategy, confirming that the introduction of the reactive component produces comparable search efficiency across simulated and real environments. Notably, both values represent substantial improvements over Fast-Cognitive's field performance, which exhibited ratios between 6.0 and 8.0 under similar conditions, representing a reduction between 48\% and 61\%.

These results confirm that the hybrid strategy transfers effectively from simulation to field deployment without parameter tuning, revealing practical applicability for autonomous OSL missions in unstructured aquatic environments.

\section{Conclusions}\label{sec6}

This work presented a novel OSL strategy combining cognitive elements from information theory with bio-inspired reactive mechanisms in a hybrid Fast-Cognitive strategy that integrates chemically reactive behavior into belief-dependent motion planning. The proposed approach triggers exploratory or exploitative actions based on real-time chemical contact rather than belief uncertainty alone, prioritizing direct source localisation over information gains.

Simulation results revealed substantial improvements over Fast-Cognitive across all metrics. Success rate increased from 87\% to 98\% overall, where reduced belief dependency enables recovery from estimation errors that cause purely cognitive methods to fail. Traveled distance decreased by 26\% and distance ratio improved from 4.21 to 3.19, confirming that detection-triggered switching reduces over-exploration while maintaining search reliability. The most meaningful results are observed in higher complex scenarios,  with a reduction of the total search approximately to 60\% when compared with Fast-Cognitive. The hybrid mechanism introduces no computational overhead, preserving Fast-Cognitive's efficiency advantages over traditional cognitive approaches. 

Field experiments in a challenging Mondego River environment validated these findings under real-world conditions. The ASV-based platform achieved 86\% success rate with an average localisation error of 3.17~m, operating in dynamic flow conditions with an upstream dam causing disturbances. The field distance ratio of 3.12 closely matched simulation predictions (3.19) and represented a 50\% improvement over Fast-Cognitive's field performance. Behavioral switching patterns observed in field trajectories confirmed that the reactive mechanism operates as designed, producing alternating exploitation-exploration dynamics consistent with simulation results. These findings show successful transition from simulation to unstructured aquatic environments without parameter tuning.

Despite these advances, limitations remain. The binary switching mechanism can produce counterproductive behavior changes under highly intermittent conditions, as observed in the single failed field experiment. Additionally, premature belief convergence cannot be corrected once declaration behavior is triggered.  Future work will address these limitations through probabilistic behavior weighting based on detection history and spatial mapping of concentrations. The introduction of Reinforcement Learning methods to improve behaviour switching and activation is also an intended path. Extension to heterogeneous multi-robot teams represents a promising direction for large-scale environmental monitoring applications. The use of multi-modal information combining chemical readings with vision-based systems is another potential path to increase the reliability of the search, by reducing divergence of the source belief.

This hybrid approach establishes a practical framework for autonomous chemical source localisation, combining bio-inspired reactivity with belief-driven adaptability while maintaining computational efficiency suitable for resource-constrained robotic platforms.

\bibliographystyle{unsrt}
\bibliography{references}  

\end{document}